\documentclass[sigconf]{acmart}

\renewcommand\footnotetextcopyrightpermission[1]{} 
\usepackage{booktabs} 
\usepackage{hyperref}
\usepackage{algorithm}
\usepackage{algpseudocode}
\usepackage{amsmath}
\usepackage{amssymb}
\usepackage[utf8]{inputenc} 
\usepackage[T1]{fontenc}    
\usepackage{hyperref}       
\usepackage{url}            
\usepackage{booktabs}       
\usepackage{amsfonts,amsmath,amssymb}       
\usepackage{nicefrac}       
\usepackage{microtype}      
\usepackage{graphicx}
\usepackage{multirow}
\usepackage{subcaption}

\begin{document}
\title{Credit Card Fraud Detection in e-Commerce: An Outlier Detection Approach}

\author{Utkarsh Porwal}
\affiliation{%
  \institution{eBay Inc}
  \city{San Jose}
  \state{California}
}
\email{uporwal@ebay.com}

\author{Smruthi Mukund}
\authornote{This work was done while the author worked at eBay}
\affiliation{%
  \institution{eBay Inc}
  \city{San Jose}
  \state{California}
}
\email{smruthi.mukund@gmail.com}

\begin{abstract}
Often the challenge associated with tasks like fraud and spam detection is the lack of all likely patterns needed to train suitable supervised learning models. This problem accentuates when the fraudulent patterns are not only scarce, they also change over time. Change in fraudulent pattern is because fraudsters continue to innovate novel ways to circumvent measures put in place to prevent fraud. Limited data and continuously changing patterns makes learning significantly difficult. We hypothesize that good behavior does not change with time and data points representing good behavior have consistent spatial signature under different groupings. Based on this hypothesis we are proposing an approach that detects outliers in large data sets by assigning a consistency score to each data point using an ensemble of clustering methods. Our main contribution is proposing a novel method that can detect outliers in large datasets and is robust to changing patterns. We also argue that area under the ROC curve, although a commonly used metric to evaluate outlier detection methods is not the right metric. Since outlier detection problems have a skewed distribution of classes, precision-recall curves are better suited  because precision compares false positives to true positives (outliers) rather than true negatives (inliers) and therefore is not affected by the problem of class imbalance. We show empirically that area under the precision-recall curve is a better than ROC as an evaluation metric. The proposed approach is tested on the modified version of the Landsat satellite dataset, the modified version of the ann-thyroid dataset and a large real world credit card fraud detection dataset available through Kaggle where we show significant improvement over the baseline methods.
\end{abstract}


\maketitle

\section{Introduction}
Outlier detection is an important problem with several applications. The goal in outlier detection is to find those data points that contain useful information on abnormal behavior of the system described by the data. Such data points are a small percentage of the total population and identifying and understanding them accurately is critical for the health of the system. 

Credit card fraud detection is one such problem that is often formulated as an outlier detection problem. Credit card fraud is one of the common type of frauds that occur in e-commerce marketplaces and it is important to have robust mechanisms in place to detect it in timely manner. In credit card fraud detection, the situation aggravates by the fact that the fraudulent behavior patterns keep changing. This is because fraudsters keep innovating novel ways to scam people and online systems. Combination of changing patterns and fewer labeled data points makes it an extremely challenging problem to keep any marketplace safe and secure.

In this work we flip the problem and rather than focusing on fraudulent behavior we try to learn good behavior. We hypothesize that unlike bad behavior good behavior does not change with time and data points representing the good behavior have consistent spatial arrangements under different groupings. Based on this hypothesis we propose an ensemble of clustering methods for outlier detection. We create an ensemble by running clustering with different clustering parameters such as by varying \textit{k} in \textit{k}-means. Each data point is assigned to a cluster in each run and therefore after \textit{k} runs, every data point can be represented by \textit{k} different centroids (corresponding to their assigned cluster of each run). These centroids vectors can be considered as different fingerprints of the data point and can be combined to generate one signature representation of the corresponding data point. Signature of the data sample can represent different properties of that data sample depending on how it is generated. In our case we are interested in the signature that represents the good behavior or consistency as per our hypothesis.  We generate this signature by combining these \textit{k} centroids per data point to generate a single score per data point weighted by the size of their respective cluster. This weighted similarity score represents consistency signature and measures consistency of individual data point or good behavior. We hypothesize that low consistency can be seen as a sign of outlier-ness. The main advantages of our method are (i) No prior knowledge of outliers or inliers is needed. (ii) The proposed algorithm is easy to scale as it can easily be implemented in a distributed manner. (iii) Proposed algorithm is general in nature and does not require \textit{k}-means algorithm as the only base clustering algorithm. One can use any clustering algorithm as long as it assigns a cluster to every data point and the said cluster can be represented as a vector in the feature space. Experimenting with different clustering algorithms can be part of the future work. 

Additionally, we also argue for a better evaluation metric for outlier detection techniques. Commonly used area under the Receiver Operating Characteristic curve is not the right metric because inliers (true negatives) are significantly higher than outliers (true positives). Precision-recall curves are better suited  because precision compares false positives to true positives (outliers) rather than true negatives (inliers) and therefore is not affected by the problem of class imbalance.

The performance of our approach is tested on three publicly available datasets - the modified version of the Landsat satellite dataset of UCI machine learning repository, the modified version of the ann-thyroid dataset of the UCI machine learning repository and the credit card fraud detection dataset available in Kaggle \cite{dal2015calibrating}. 

The rest of the paper is organized as follows. Section~\ref{sec:related} describes the related research in the area of outlier detection. Section~\ref{sec:hypo} explains our hypothesis. Section~\ref{sec:cd} outlines the algorithm for consistency estimation. Data sets descriptions and results are outlined in Section~\ref{sec:exp}. Section~\ref{sec:observation} highlights the observations followed by conclusion in Section~\ref{sec:conclusion}.

\section{Related Work}
\label{sec:related}
Several approaches have been proposed for outlier detection based on different assumptions and techniques. One of the most common and intuitive ways to detect outliers is by assuming an underlying distribution for the data. For Gaussian distribution we can either model all the attributes together as one multivariate distribution or by modeling each attribute as a separate Gaussian distribution \cite{grubbs1969procedures}. 
Other commonly used approaches are based on the distance measure. In this approach it is assumed that all regular data points lie close to each other and outliers are far from them \cite{Tan:2005}. Nearest neighbor techniques have been employed to detect outliers with this assumption \cite{ramaswamy2000efficient}. 
Another approach is clustering based approach where it is assumed that regular data points make clusters and anomalies are either not part of any cluster or make separate clusters \cite{Tan:2005}. However, data often makes different cluster for different set of attributes and these clusters lie in different subspaces. To overcome this issue, sub space clustering is performed with an assumption that all the subspaces are axis parallel to reduce the complexity of exploring subspaces. Lazarevic et al. \cite{Lazarevic} proposed an ensemble method where ensemble was created using different set of features. They hypothesized that each member of the ensemble identifies different outliers because of different features and their scores are combined to produce a common score to identify outliers. Our approach is similar to this approach with a different way of creating the ensemble and combining their score. Chandola et al. \cite{Chandola:2009} and Campos et al. \cite{Campos} cover a more comprehensive analysis of related work in the field of outlier detection. Recently ensemble of unsupervised methods is used for outlier detection \cite{aggarwal2015theoretical} \cite{satimage2} \cite{dal2015calibrating}. Ensemble is proven to be more effective than a single method and is heavily used in supervised setting \cite{dietterich2000ensemble} \cite{porwal2012ensemble}. However, use of ensemble in unsupervised setting and for outlier detection has its own challenges and is not studied as thoroughly as the use of ensemble for supervised setting. Aggarwal et al. \cite{aggarwal2015theoretical} and Zimek et al. \cite{zimek2014ensembles} did separate studies to consolidate the work done using ensembles for outlier detection and highlighted the challenges associated with it.   

\section{Hypothesis}
\label{sec:hypo}
Consider an application of credit card fraud detection. Here, it is easier to obtain samples with good non-fraud behavior than samples that exhibit a fraudulent pattern as the latter is scarce and time variant. Once a fraud pattern is accurately determined it is just a matter of time before fraud shifts to a different area exhibiting a totally different pattern. However, if we can develop a method that can estimate a measure of consistent behavior (good behavior) for each data point then we can identify outliers as data points with low consistency score. We refer to the data points that exhibit good non-fraud patterns as consistent data points. We consider data points that belong to the same cluster (or close proximity clusters) under different spatial groupings as being consistent. Based on this understanding, we attempt the problem of outlier detection by estimating a consistency score for every data point\cite{porwal2017outlier}. One way to estimate the consistency for every data point is by running an ensemble of unsupervised clusters. Since the spatial arrangements of consistent points do not change, they should form closed clusters. We run the simple \textit{k}-means clustering algorithm \textit{N} times for different values of \textit{k} where \textit{k} can range from 2 to \textit{K}. The \textit{N} runs of \textit{k}-means on the sample set will assign each data point to \textit{N} clusters with \textit{N} centroids associated with their respective clusters. Note that each data point is represented by  \textit{N} centroids corresponding to the \textit{N} runs. 

Next step is to estimate the consistency score for each data point. We use dot product to calculate this consistency score. Dot products are often used as a proxy for similarity of two vectors. However we have several vectors (centroids) per data point (from different runs of \textit{k} means) representing different finger prints of the corresponding data point. We have to calculate one signature from these different finger prints in the form of vectors. One way is to compute how similar all these finger prints are to each other and somehow combine them to generate one scalar signature (consistency score of that data point). We do this by adding dot products of all $\binom N2$ combinations of centroids for every data point. We have to normalize this score to get a score between 0 and 1 per data point where 1 would mean highly consistent and 0 would mean poor consistency. However, rather than adding the dot products of all centroid combinations per data point we propose taking weighted average of their dot products based on the size of the cluster. The motivation behind this is that although inliers and outliers may fall in different clusters under different groupings but they still are found together within their respective classes. In other words, if outliers are clubbed together under different groupings, size of those clusters will be small. Therefore, a size weighted score will be a way to discount consistency of outlier samples. 
  
\begin{algorithm}{}\label{alg:algo}
\caption{Consistency Estimation Method}
\begin{algorithmic}

\Require 
	\State $\Theta \leftarrow$ dataset with $n$ data points
	\State AvgSimScore $\leftarrow n$ dimensional array
\Procedure{}{}
	\State Run $k$-means on $\Theta$ for $K$ = \{$k_1$,$k_2$,$k_3$ .. $k_N$\}
	\For{$i=1...n$}
		\State $\mathbf{x_\emph{i}} \sim \Theta$ = $Set(\mathbf{c_1},\mathbf{c_2}..\mathbf{c_N})$, where $\mathbf{c_k}$ is a centroid of a cluster $\mathbf{x}$ belongs to
                \State AvgSimScore($i$) =$weighted\_score$
	\EndFor

\EndProcedure
\end{algorithmic}
\label{alg:cds}
\end{algorithm}

\section{Consistency Estimation}
\label{sec:cd}
Given a set of data points \{$\Theta$ = $\textbf{x}_1$,$\textbf{x}_2$,$\textbf{x}_3$ .. $\textbf{x}_n$\} that contain a relatively small sample of outliers  \{$O$ = $\textbf{o}_1$,$\textbf{o}_2$,$\textbf{o}_3$ .. $\textbf{o}_m$\} such that $m \ll n$, the goal is to effectively identify the outlier pool. Each data point can have multiple attributes and each attribute is denoted by $x_{ij}$ where $i$ is the data point index and $j$ is the attribute index. A consistent data point set is determined using an ensemble of $k$-means clusters, $K$ = \{$k_1$,$k_2$,$k_3$ .. $k_N$\}. The value of $k$ ranges from a small number to a large number where the largest run $k_N \le n$, the sample size. Each data point, after running the suit of $k$-means clusters, will belong to $N$ clusters with $C$ centroids where $C$ = \{$\textbf{c}_1$,$\textbf{c}_2$, .. ,$\textbf{c}_N$\}.

We measure the similarity of two centroids based on their vector similarity scores. \textbf{For each data point}, consistency is determined by estimating the weighted average similarity score(s) of the centroids it belongs to as shown below


\begin{displaymath}{weighted\_score = \frac{\sum_{i=1}^{N-1}\sum_{j=i+1}^{N} (n_i + n_j)cos(C_i,C_j)}{\sum_{i=1}^{N-1}\sum_{j=i+1}^{N} (n_i + n_j)}}\end{displaymath} 

where $cos$ is the cosine similarity metric and $C_i$ is the vector centroid of cluster $i$. Here $n_i$ and $n_j$ are the number of data points in the $i$ and $j$ clusters. If the average similarity score for a data point is very high (closer to 1) then the centroids are very close to each other, and the data point is considered to be consistent. Algorithm \ref{alg:cds} highlights all the steps of consistent data selection method.
\begin{figure}
\centering
\includegraphics[height=3in, width=3in]{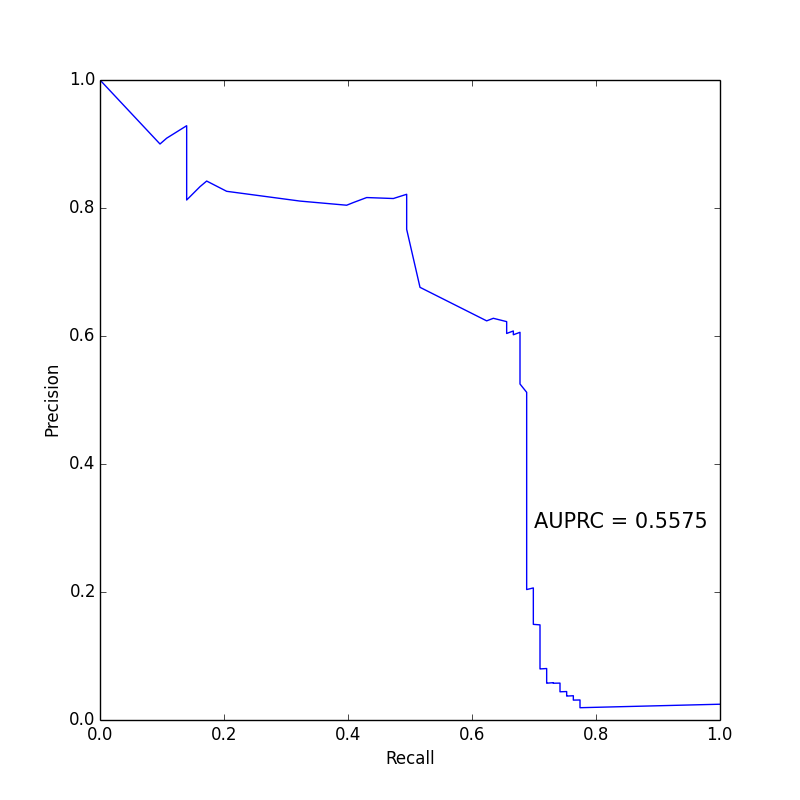}
\caption{Precision Recall curve of the proposed method on Thyroid dataset}
\label{fig:thypr}
\end{figure}
\section{Experiments}
\label{sec:exp}
The motivation for this approach comes from trying to identify fraudulent consumers on an e-Commerce platform. On a data set that contains transactions for a given day, identifying fraudulent patterns is not easy. Each time the e-Commerce company introduces new consumer aided features or imposes restrictions on certain transactional behaviors, it opens new doors and avenues for some consumers to misuse and abuse the platform. Our algorithm shows tremendous potential in identifying fraudulent transactions. We observed that the samples in the lower score buckets, had higher recall for fraudulent transactions. However, due to the confidentiality of the data set, these results cannot be reported in this paper. In order to show the efficacy of the proposed approach we test it on publicly available datasets.  We first show a proof of concept on two data sets from the UCI machine learning repository. Finally we run a thorough analysis on a credit card fraud detection dataset available in Kaggle\cite{dal2015calibrating}\footnote{\href{url}{https://www.kaggle.com/mlg-ulb/creditcardfraud}}.

\begin{figure}
\centering
\includegraphics[height=3in, width=3in]{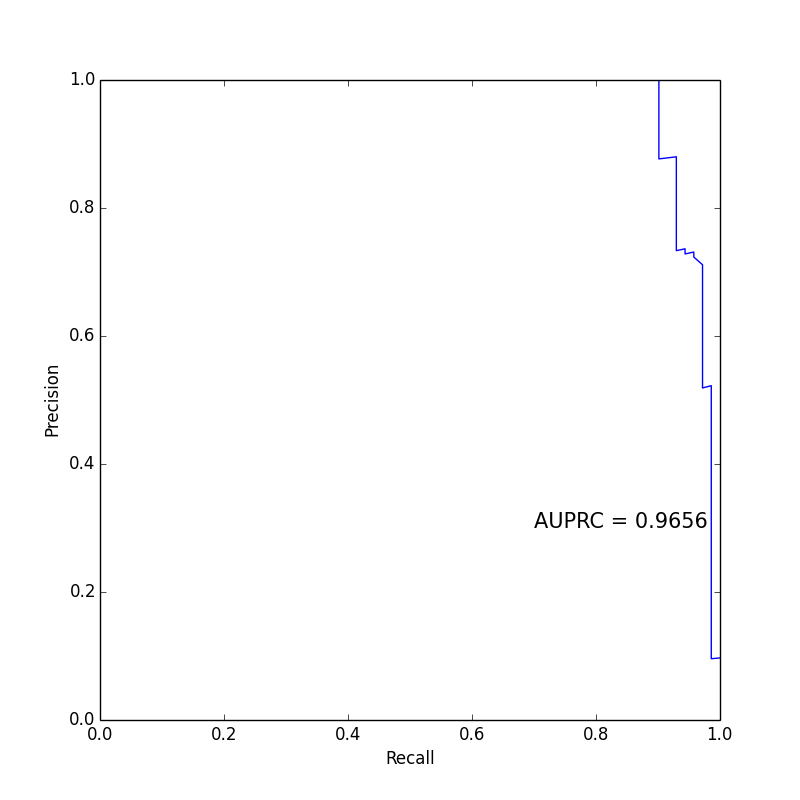}
\caption{Precision Recall curve of the proposed method on Satimage-2 dataset}
\label{fig:satpr}
\end{figure}

\subsection {Implementation Details}

We run the proposed algorithm for finding the consistent data points and get a weighted similarity score as consistency score for each data point. Our hypothesis is that the most consistent data points will have a high consistency score. We also propose that the data points with poor consistency scores are more likely to be outliers. Therefore, we flag the ones with poor score as fraudulent transactions. However, since our algorithm is an ensemble of \textit{k}-means for different values of \textit{k}, it runs into same issues that \textit{k}-means algorithm does. It can potentially run into local minima and can be sensitive to the order in the data is presented. Therefore, we randomize our datasets and run our algorithm 10 times to get an estimate of variance these issues can incur. We report standard deviation in each experiment to quantify this variance. We also normalize datasets while running experiments along the feature dimensions as part of the preprocessing. Lastly throughout our experiments and analysis we will refer outliers as positives and inliers as negatives.

\subsection{How to select \it{K}?}
We use an ensemble of \textit{k}-means for different values of \textit{k}. It is common practice to learn ensemble of weak learners so that collectively they can learn a good hypothesis function \cite{dietterich2000ensemble}. We hypothesize that \textit{k}-means is a weak learner  because of its sensitivity to the value of \textit{k}, and having an ensemble of such learners discounts the uncertainty caused by \textit{k}-means. In our experiments we found that incrementally increasing \textit{k} with a fixed \textit{step} works just as well as the ensemble created by carefully selecting \textit{k} using a principled approach such as Silhouette Score\cite{silhouettes}. Note that Silhouette Score is expensive as the computational complexity is $O(mn^{2})$ where \textit{m} is number of features and \textit{n} is the number of data points. Therefore it could be computationally prohibiting to use Silhouette Score if the data size is big.    

\subsection{Computation Complexity}
We used an implementation of Lloyd's algorithm\cite{lloyd} which has an average run time complexity of $\mathcal{O}(knmi)$ where \textit{n} is the number of samples, \textit{m} is the number of features and \textit{i} is the number of iterations before convergence. In our proposed method we are creating an ensemble of $N$ runs and we run our algorithm $t$ times to estimate the variance ($t$ is set to 10 for all our experiments) giving it a total complexity of $\mathcal{O}(tNknmi)$. Once the ensemble is created, run time complexity of estimating the consistency score per data point is $\mathcal{O}(N^2m)$ because we have to calculate cosine similarity of $\binom N2$ combinations and cosine similarity itself is linear. Since we have to do this for every data point, complexity of estimating consistency score for all data points will be $\mathcal{O}(N^2nm)$. Therefore, total run time complexity of our algorithm is $\mathcal{O}(tNknmi) + \mathcal{O}(N^2nm)$. However, since each of the $tN$ runs in creating the ensemble are independent of each other and can be parallelized, the effective run time complexity of our algorithm is $\mathcal{O}(knmi) + \mathcal{O}(N^2nm)$. Our algorithm is used to make recommendations for further investigation, it is run as an offline job. This job needs to run as frequently as the team needs to make new recommendations. Because of the offline nature of our application, accuracy of the algorithm takes precedence over the run time complexity of the algorithm for us.

\subsection{Evaluation}
We have two classes in our application namely fraudulent and non-fraudulent transactions.  It is common to use Receiver Operator Characteristic (ROC) curves to evaluate binary classifiers as they reflect how number of correctly labeled positive samples vary with the number of incorrectly labeled negative examples. However, in our case, we have a problem of huge class imbalance as the number of samples for positive class in the credit card fraud detection dataset are only 0.172\% of the total samples. Therefore, we care more about the performance of the proposed algorithm in identifying the positive samples than in identifying the negative ones. Davis et al.\cite{davis2006relationship} argue that ROC curves can present a misleading overview of the performance if the dataset has class imbalance issue. They suggest that Precision-Recall curves are better suited for problems with large skew in classes because Precision compares false positives to true positives rather than true negatives and therefore is not affected by the problem of class imbalance. Note that because of large number of true negatives in the dataset even a big change in the number of false positives only results in the small change in the false positive rate used for estimating AUROC. Hence, we are going to use area under the precision recall curve (AUPRC) as our primary metric to evaluate the performance of the proposed and baseline methods. Note that majority of the work in the outlier detection literature has used AUROC as their metric for evaluation\cite{dal2015calibrating}\cite{satimage2}\cite{aggarwal2015theoretical}\cite{lodes} but we argue that AUPRC should also be reported because algorithms that optimize the area under the ROC curve are not guaranteed to optimize the area under the PR curve \cite{davis2006relationship}. Both metrics collectively present a complete picture of the performance of the algorithm because of their focus on different classes. We provide the definition of these metrics below for sake of completion. 

\begin{displaymath}{Precision = \frac{TP}{TP+FP}}\end{displaymath} 
\begin{displaymath}{Recall = \frac{TP}{TP+FN}}\end{displaymath} 
\begin{displaymath}{True Positive Rate = \frac{TP}{TP+FN}} \end{displaymath} 
\begin{displaymath}{False Positive Rate = \frac{FP}{FP+TN}}\end{displaymath}

\begin{figure*}
\centering
\includegraphics[width=.15\textwidth]{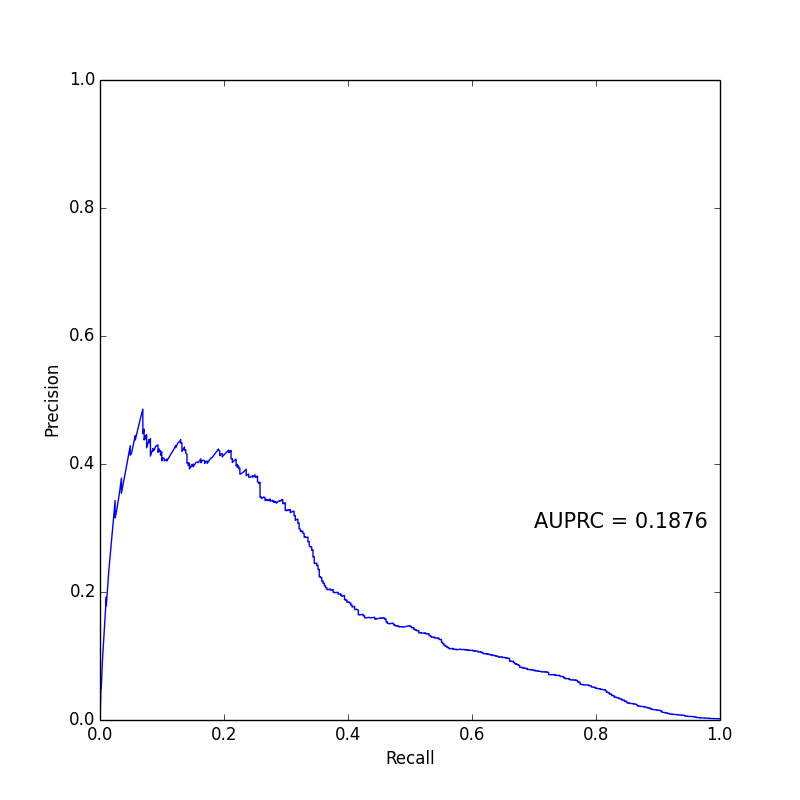}\quad
\includegraphics[width=.15\textwidth]{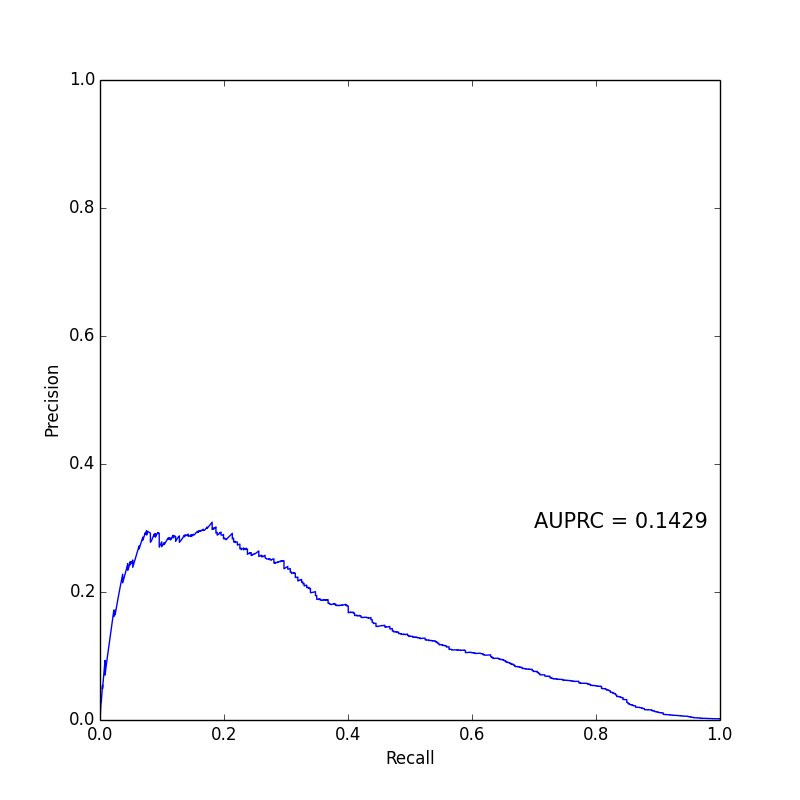}\quad
\includegraphics[width=.15\textwidth]{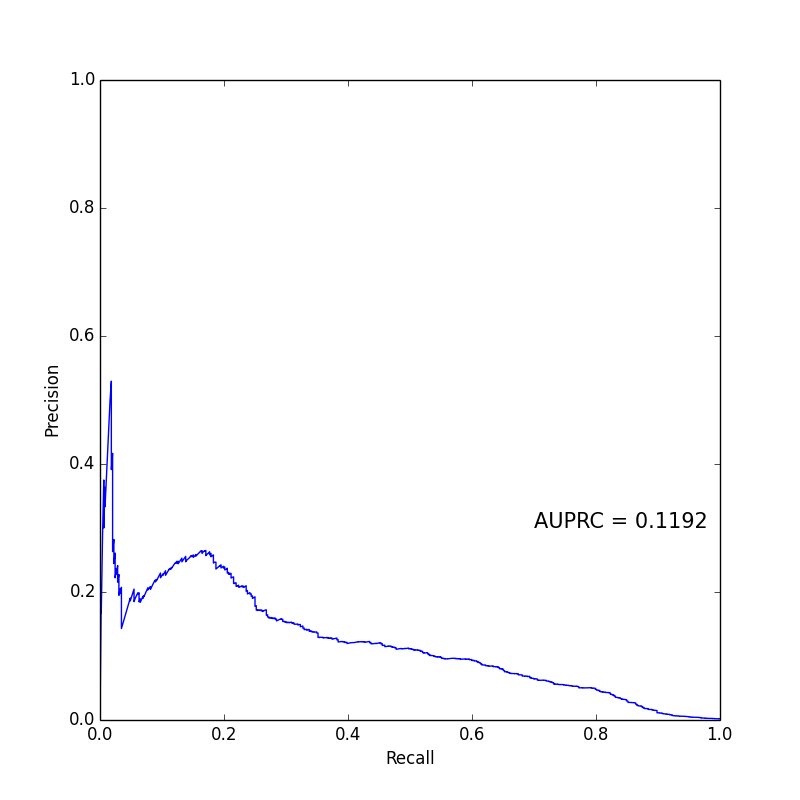}\quad
\includegraphics[width=.15\textwidth]{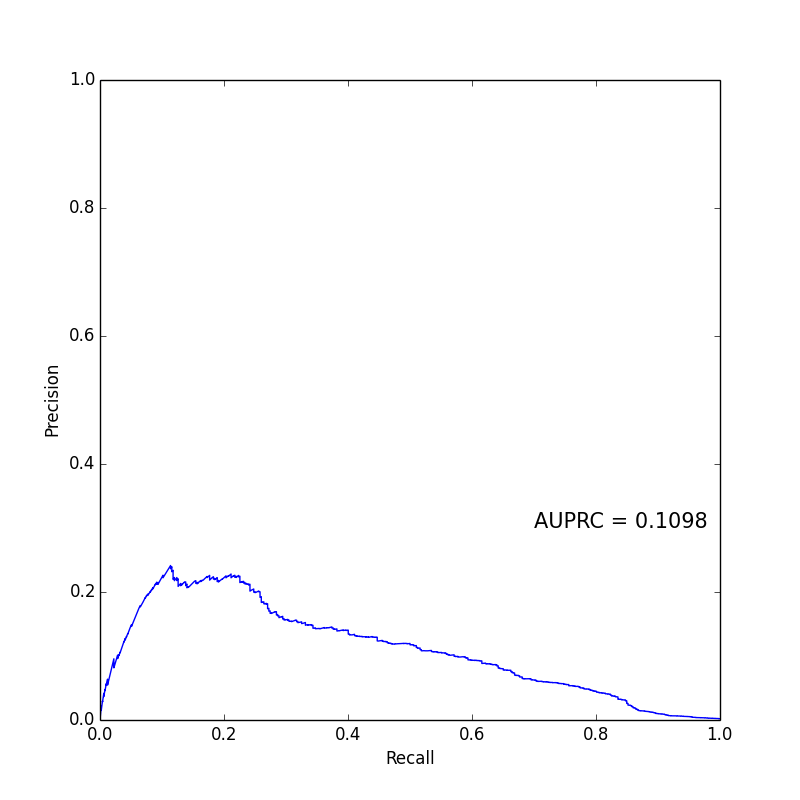}\quad
\includegraphics[width=.15\textwidth]{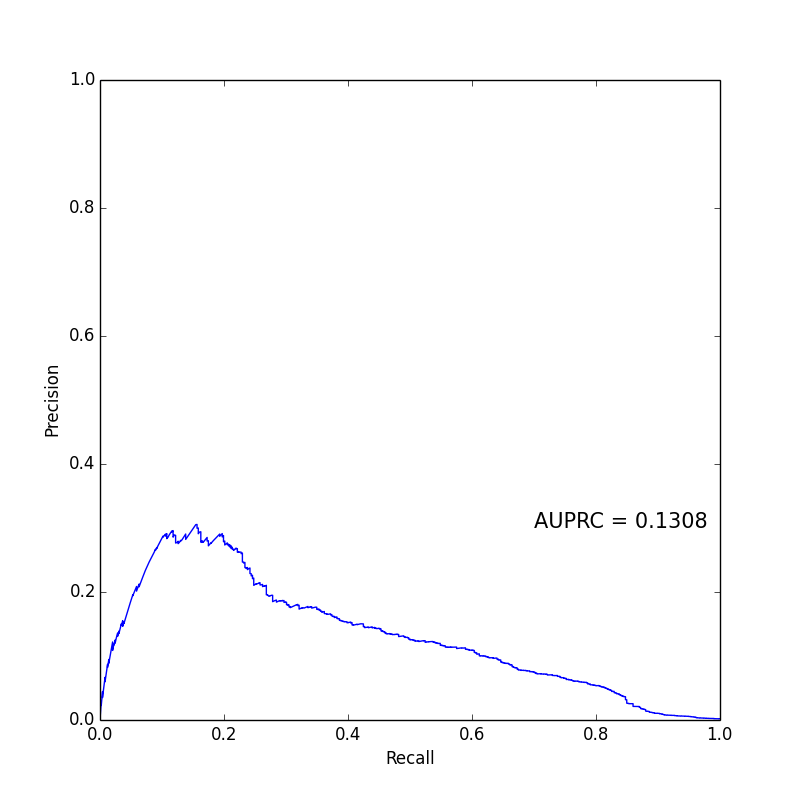}

\medskip

\includegraphics[width=.15\textwidth]{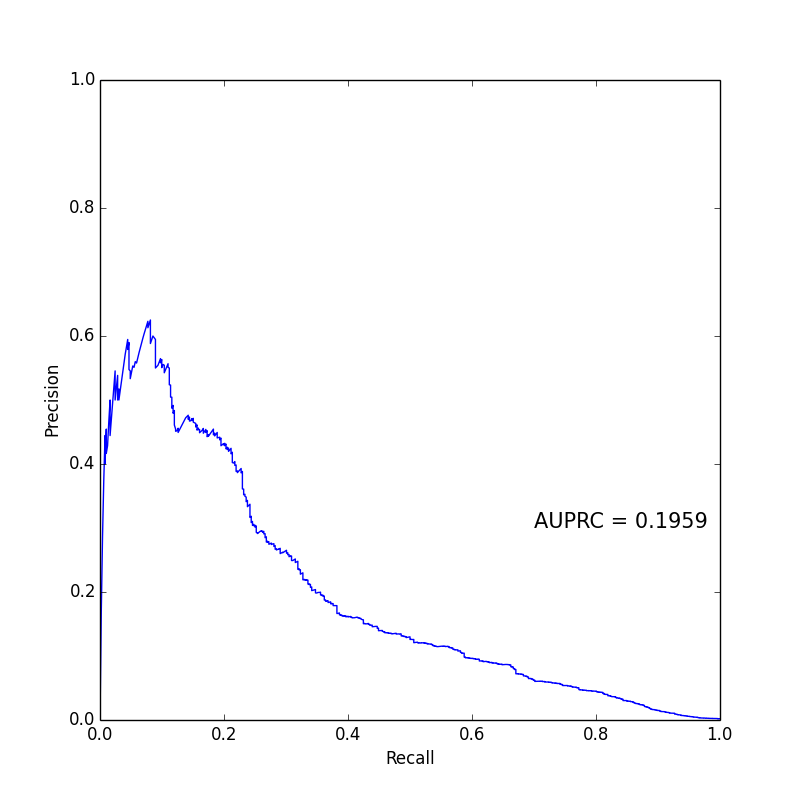}\quad
\includegraphics[width=.15\textwidth]{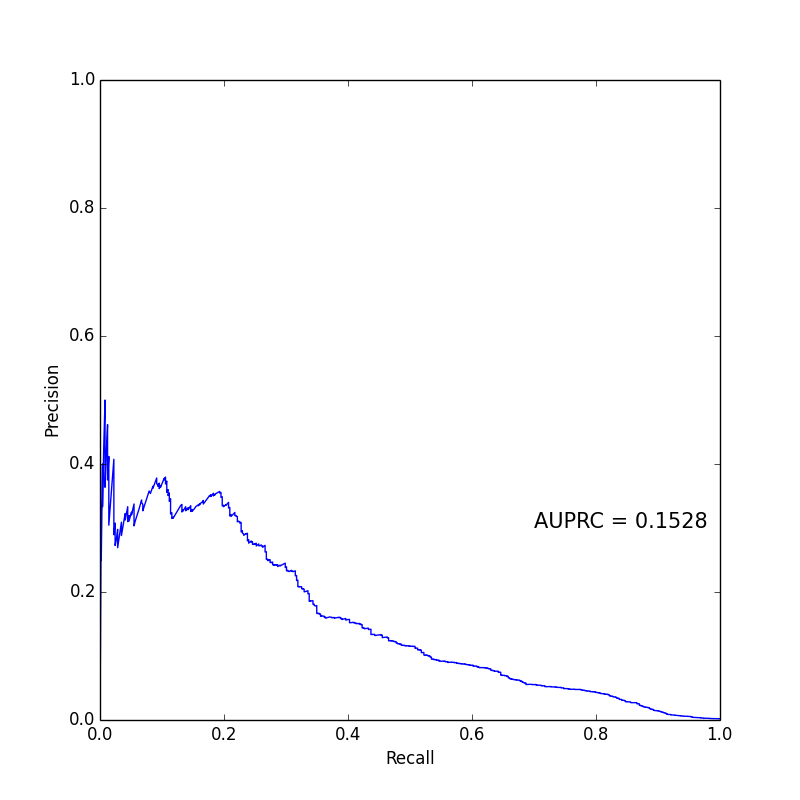}\quad
\includegraphics[width=.15\textwidth]{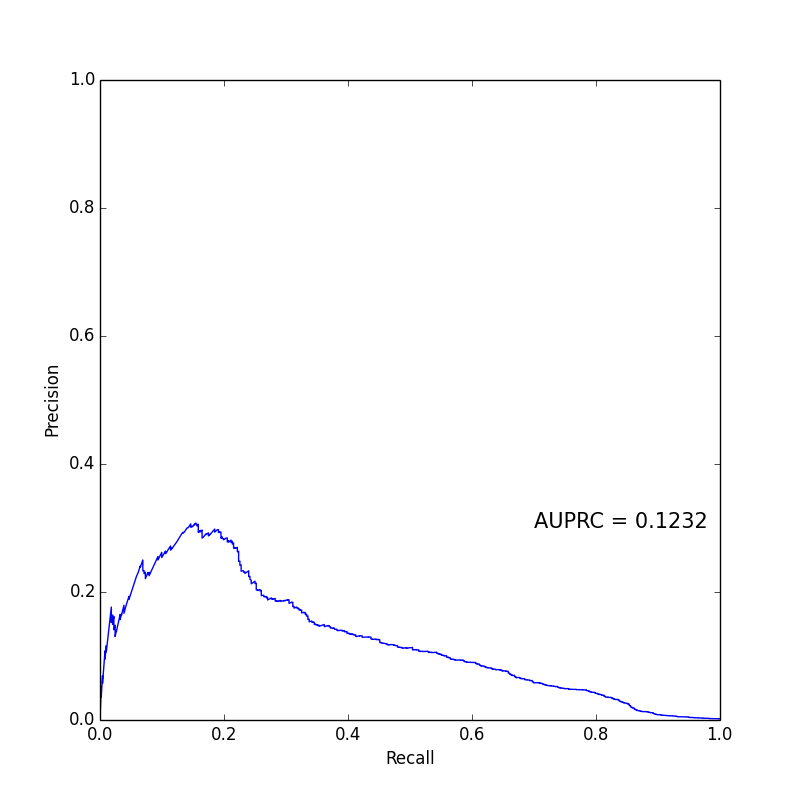}\quad
\includegraphics[width=.15\textwidth]{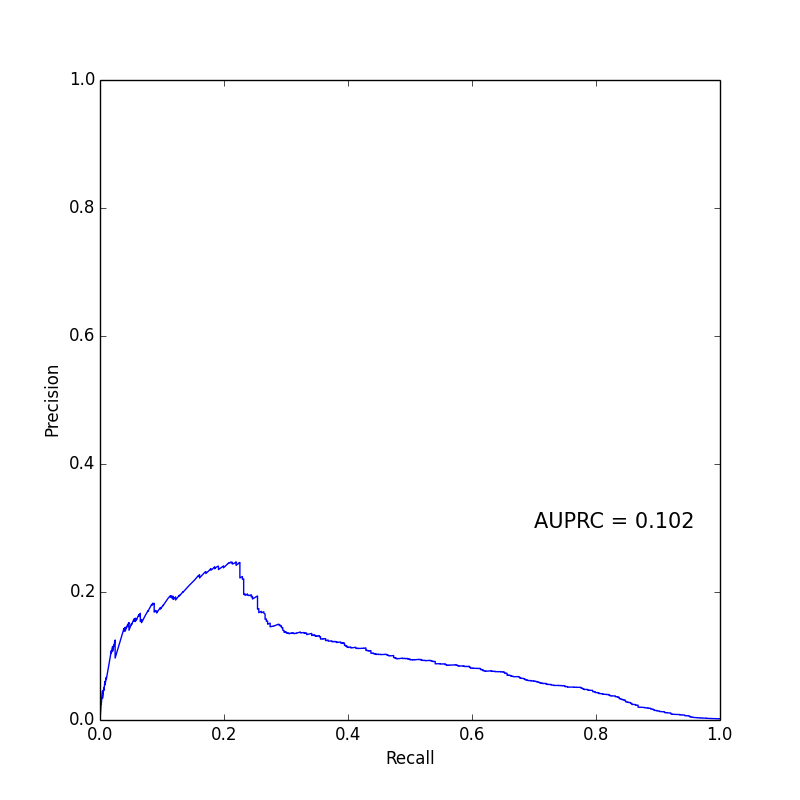}\quad
\includegraphics[width=.15\textwidth]{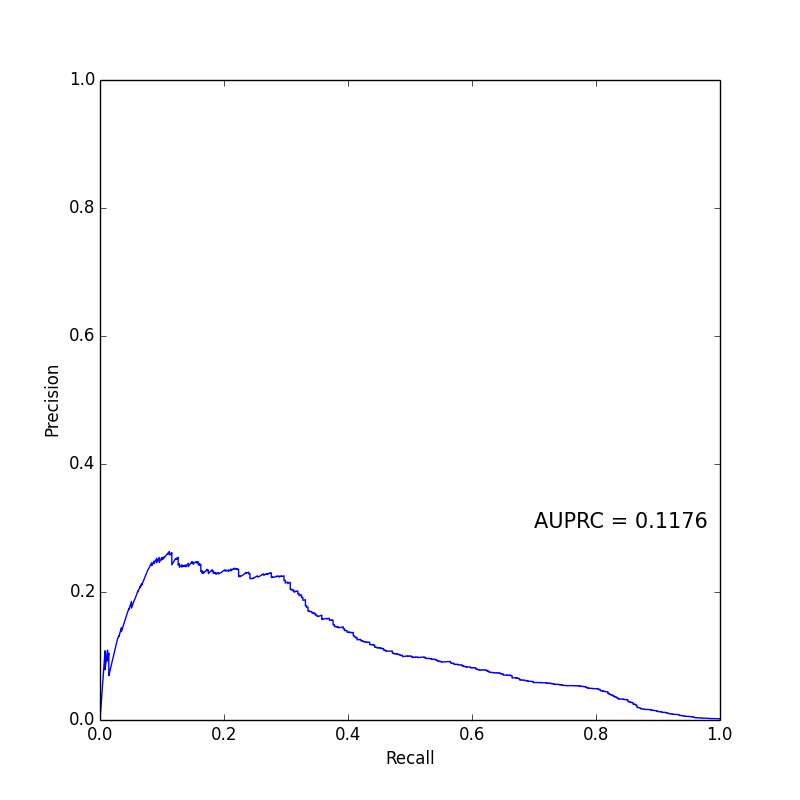}

\caption{Precision Recall curves of different runs of Isolation Forest on the Credit Card Fraud Detection Dataset}
\label{fig:prif}
\end{figure*}

\begin{figure*}
\centering
\includegraphics[width=.15\textwidth]{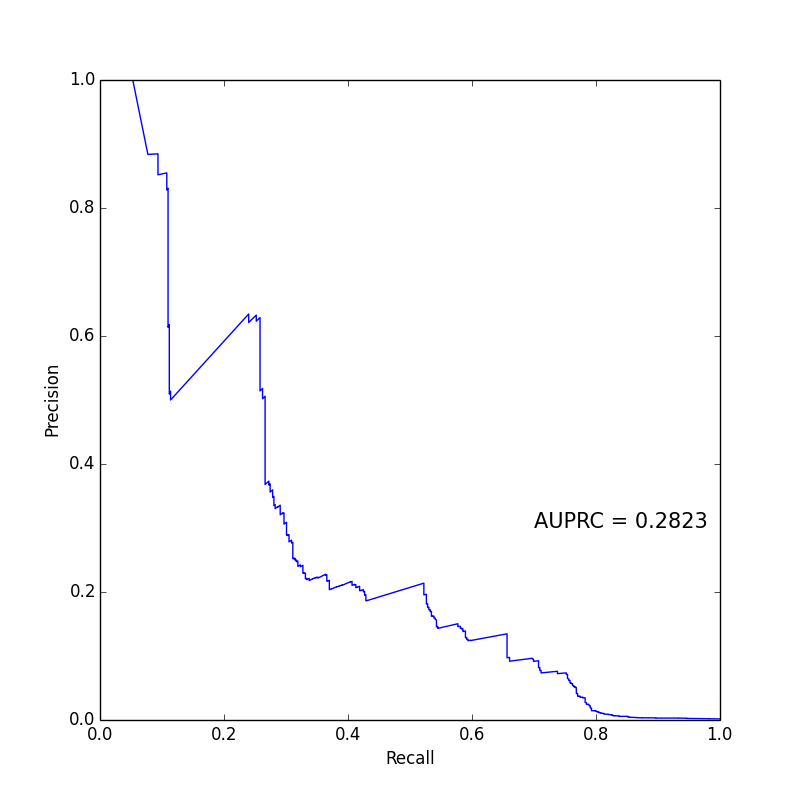}\quad
\includegraphics[width=.15\textwidth]{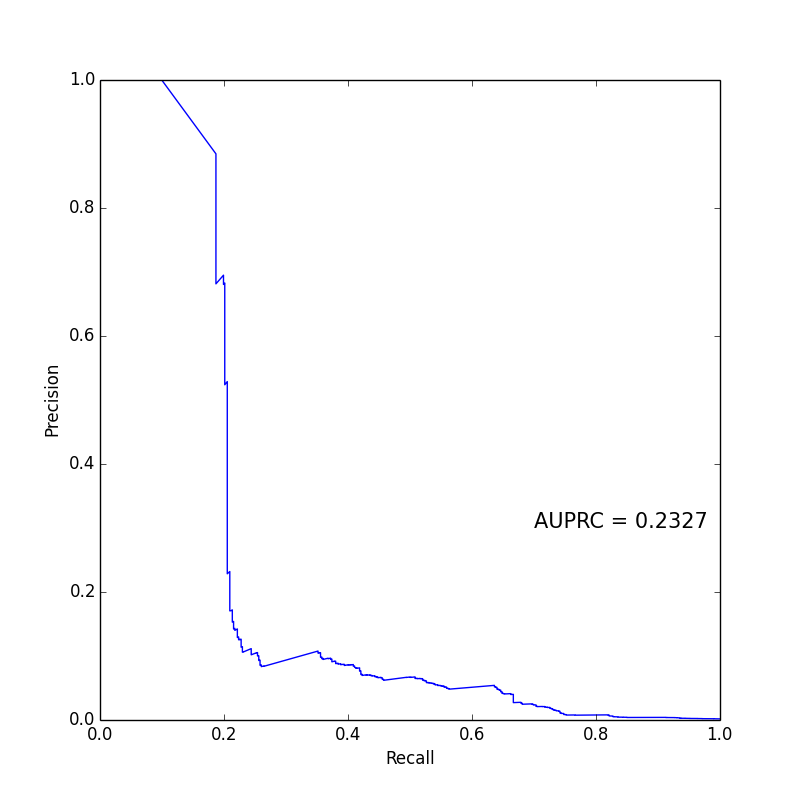}\quad
\includegraphics[width=.15\textwidth]{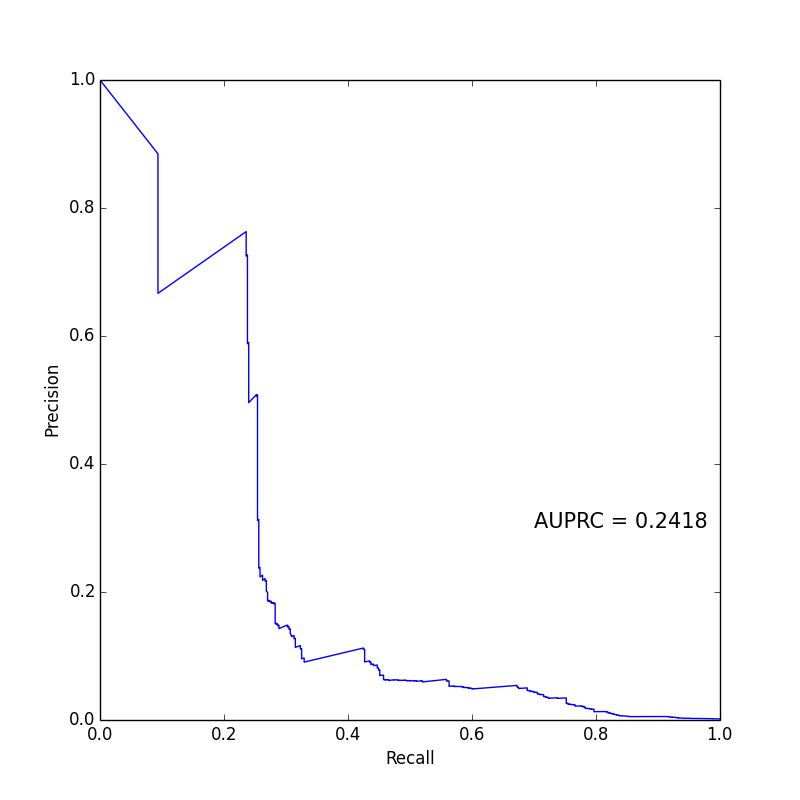}\quad
\includegraphics[width=.15\textwidth]{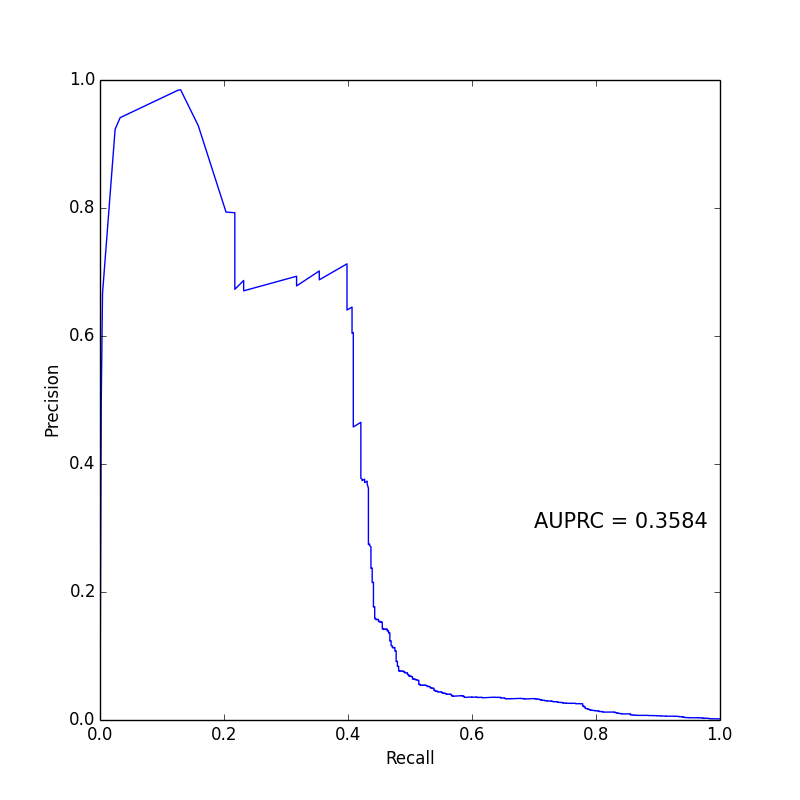}\quad
\includegraphics[width=.15\textwidth]{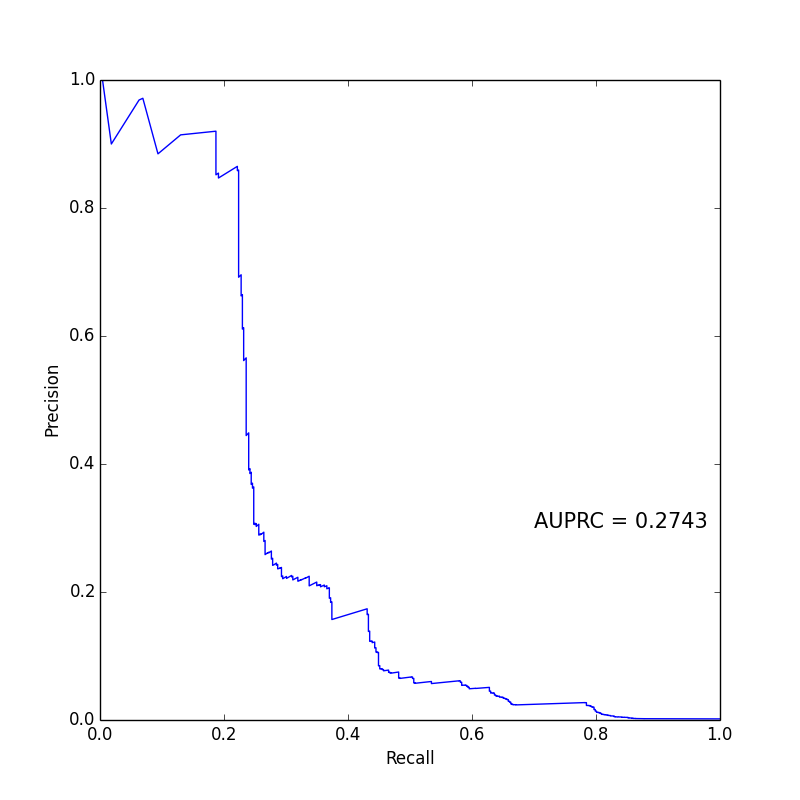}

\medskip

\includegraphics[width=.15\textwidth]{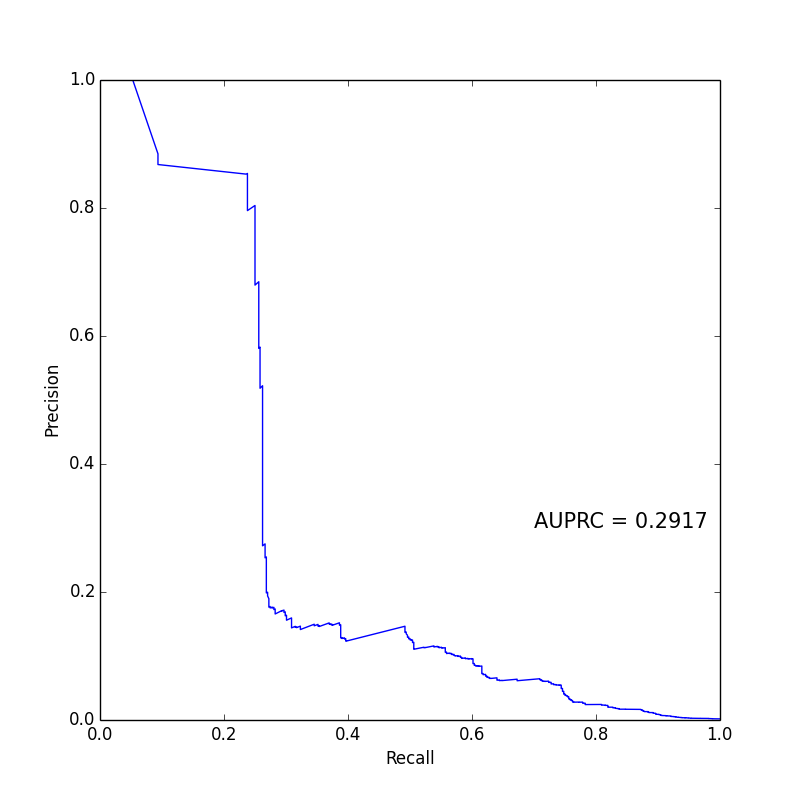}\quad
\includegraphics[width=.15\textwidth]{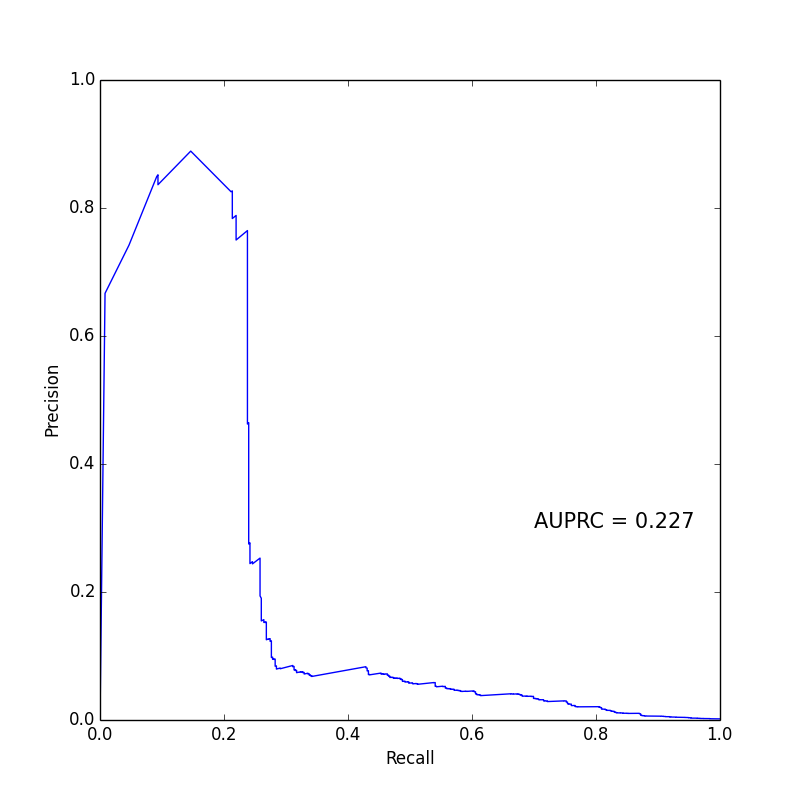}\quad
\includegraphics[width=.15\textwidth]{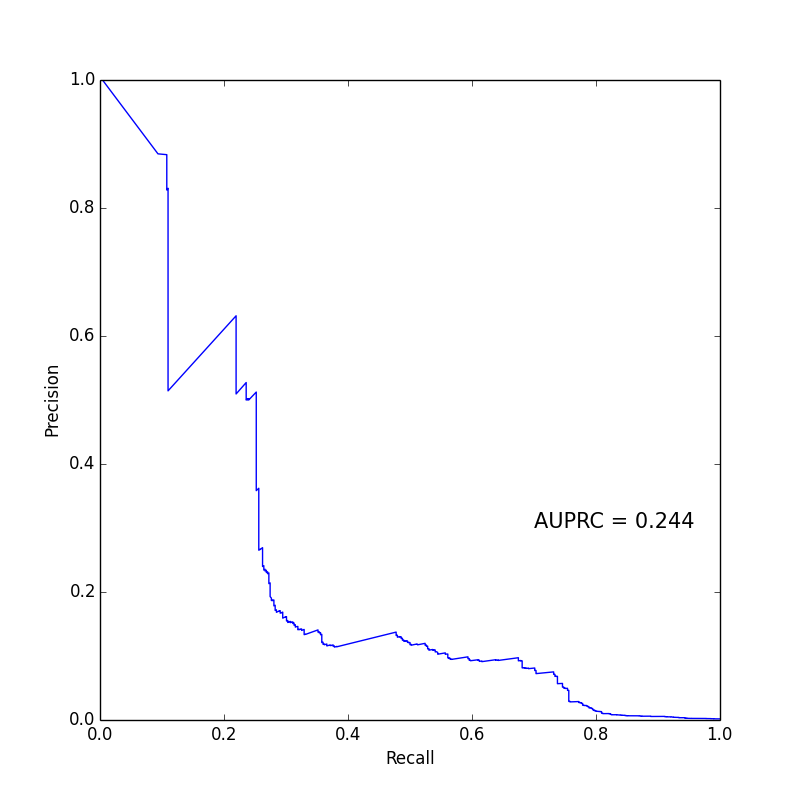}\quad
\includegraphics[width=.15\textwidth]{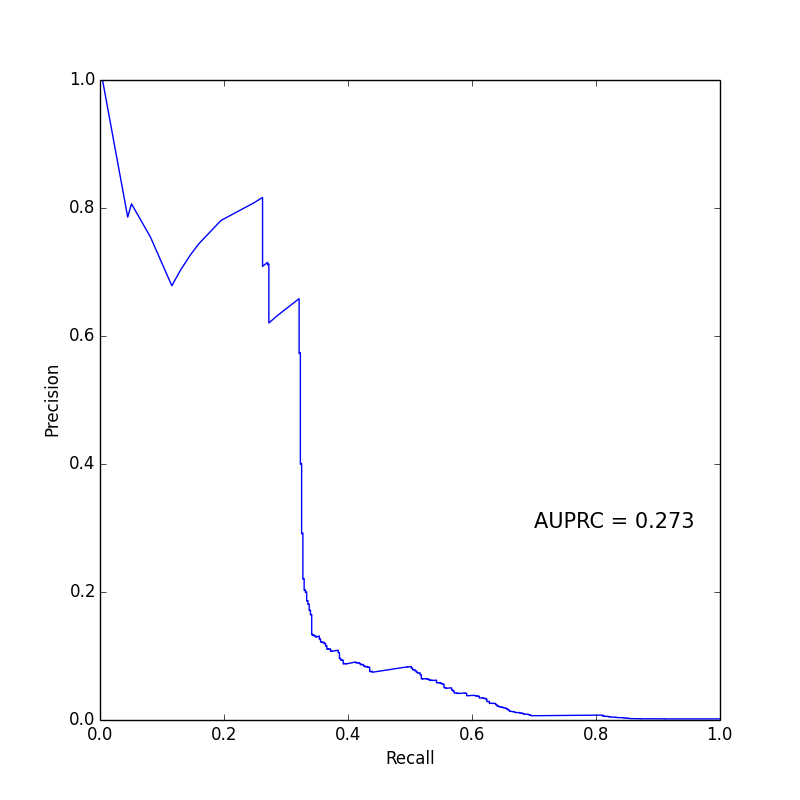}\quad
\includegraphics[width=.15\textwidth]{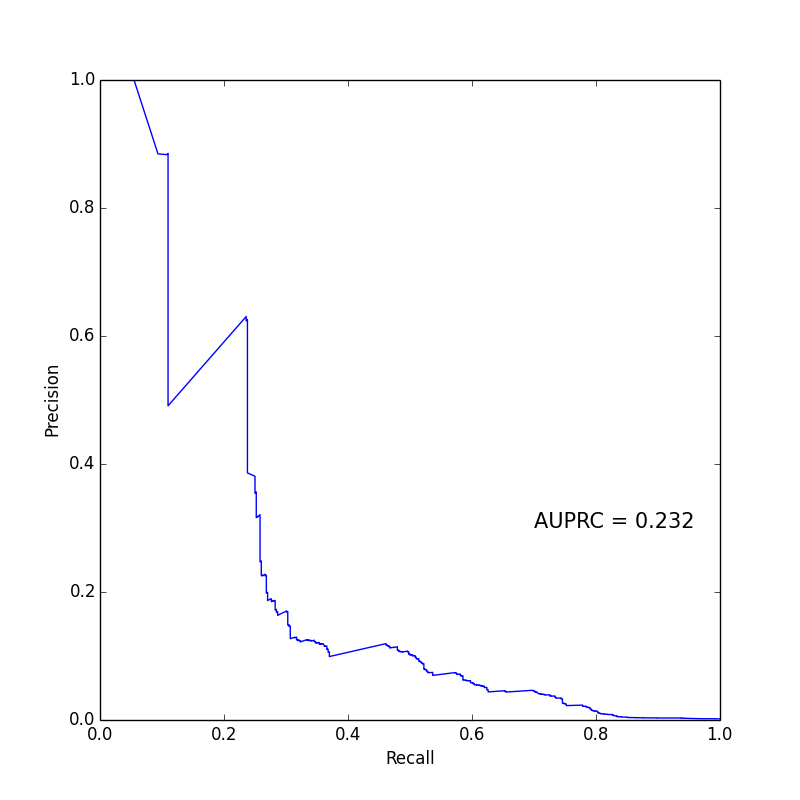}

\caption{Precision Recall curves of different runs of the proposed algorithm on the Credit Card Fraud Detection Dataset}
\label{fig:prpm}
\end{figure*}

\subsection{Satimage-2}
The Landsat satellite dataset from the UCI repository\cite{uci} is a small sub-area of a scene, consisting of 82 x 100 pixels. Each line of data corresponds to the multi-spectral values of a 3x3 square neighborhood of pixels completely contained within the 82x100 sub-area. Each line contains the pixel values in the four spectral bands (converted to ASCII) of each of the 9 pixels in the 3x3 neighborhood and a number indicating the classification label of the central pixel. The number is a code for the different types of soil. There are no examples with class 6 in this dataset.
 
The Satimage-2 dataset\footnote{\href{url}{http://odds.cs.stonybrook.edu/\#table1}} is prepared from the original dataset for the purpose of outlier detection by combining training and test data. Class 2 is down-sampled to 71 outliers, while all the other classes are combined to form an inlier class. Total number of data points in the dataset are 5803 and outliers form the 1.2\% of the total data points. Number of features in the dataset is 36.

Our proposed method gets near perfect performance on this dataset on metric AUROC with minimal standard deviation. The results are shown in the Table~\ref{tab:satimage}. 

\begin{table}[ht]
\centering
\begin{tabular}{|c|c|c|c|} \hline
\multicolumn{2}{|c|}{AUPRC}& \multicolumn{2}{c|}{AUROC} \\
\cline{1-4}
Mean & Standard Deviation &  Mean & Standard Deviation \\ \hline
0.95 & 0.0034 & 0.99 & 0.0003 \\ \hline
\end{tabular}
\caption{AUROC and AUPRC of the proposed method on the Satimage-2 dataset after 10 runs}\label{tab:satimage}
\end{table}

Zimek et al.\cite{satimage2} proposed ensemble of outlier detection techniques based on subsampling and used Satimage-2 in their work. They argued that subsampling leads to a better model than using the full data and creating an ensemble of such models created from subsampling further improves the performance. They used LOF\cite{lof}, LDOF\cite{ldof} and LoOP\cite{loop} as base methods and AUROC as their metric. Figure 7 of \cite{satimage2} shows us that only in the case of sample fraction of 0.1 their method comes close to the performance of our proposed method for all three base classifiers. Our method significantly outperforms their method in all other settings for all three base methods. Likewise, Aggarwal et al.\cite{aggarwal2015theoretical} also used an ensemble of LOF and average \textit{k}-NN based on subsampling for different \textit{k}. Figure 8 of their work\cite{aggarwal2015theoretical} shows that our proposed method is either comparable to theirs in few settings and outperforms theirs in most settings. Note that both \cite{satimage2} and \cite{aggarwal2015theoretical} do not report numbers of their performance and only provide their results in the form of figures which are not reproducible. Therefore we only could only do visual comparison with our method. Both of these works used AUROC as their metric like most of the work in literature. However, we argue that AUPRC is a better metric for outlier detection and no one reported their performance on AUPRC. We argue that performance measured on AUPRC is a better measure of the quality of any outlier detection algorithm. As shown in the Table~\ref{tab:satimage}, our performance using AUPRC is also high with minimal standard deviation. Since the performance of our method is highly consistent we show the AUPRC curve of one of the 10 runs in Figure \ref{fig:satpr}.

\subsection{Thyroid} 
\begin{table*}[ht]
\centering
\begin{tabular}{|c|c|c|c|c|c|c|} \hline
  \multicolumn{2}{|c|}{Proposed Method} & \multirow{2}{*}{LODES}& \multirow{2}{*}{HiCS}& \multirow{2}{*}{OutDST}& \multirow{2}{*}{FastABOD}& \multirow{2}{*}{LOF} \\
\cline{1-2}
 Mean & Standard Deviation & &&&&  \\ \hline
\textbf{0.7480} & 0.0037 & 0.6840  & \textbf{0.7682} & 0.512 & 0.5558 & 0.6714\\ \hline

\end{tabular}
\caption{AUROC of the proposed method after 10 runs and the other baseline methods as reported by \cite{lodes} on the Thyroid dataset}\label{tab:thyroid}
\end{table*}
The original thyroid disease (ann-thyroid) dataset from the UCI machine learning repository\cite{uci} is a classification dataset suited for training ANNs. It has 3772 training instances and 3428 testing instances. It has 15 categorical and 6 real attributes. The problem is to determine whether a patient referred to the clinic is hypothyroid.  Normal (not hypothyroid), hyper-function and subnormal functioning are the three classes in the original dataset. For the purpose of outlier detection, 3772 training instances are used, with only 6 real attributes. The hyper-function class is treated as outlier class and other two classes are inliers. Total number of outliers in the dataset are 93 i.e. 2.46\% of the total data points.

\begin{figure}
\centering
\includegraphics[height=3in, width=3in]{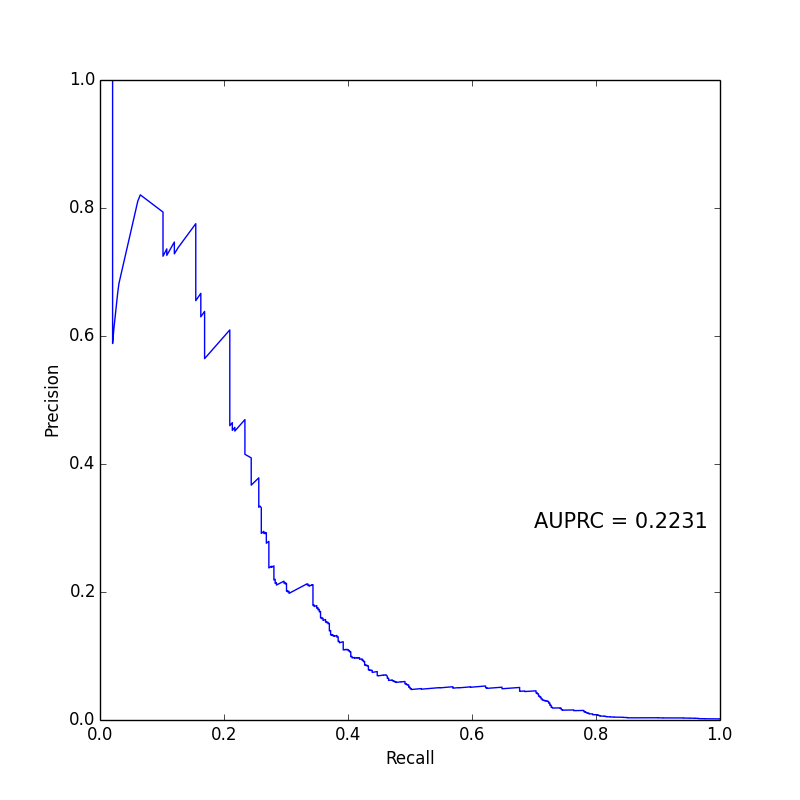}
\caption{Precision Recall curve of the proposed method on the credit card fraud detection dataset ordered by time of the transactions }
\label{fig:pr}
\vskip -15pt
\end{figure}

Sathe et al.\cite{lodes} combined spectral techniques with local density-based methods for outlier detection (LODES) and used Thyroid dataset in their work. They also reported performance of other methods such as LOF\cite{lof}, HiCS\cite{hics}, OutDST\cite{outdst} and FastABOD\cite{angle}. Performance of our proposed method after 10 runs along with the other baseline methods is shown in the Table~\ref{tab:thyroid}. We got the second-best performance out of all the method with minimal standard deviation. However, Sathe et al.\cite{lodes} did not report their performance on AUPRC which we argue to be a better metric for outlier detection. On AUPRC, we got a mean of 0.5584 with a standard deviation of 0.0081 after 10 runs of the proposed method. Note that the standard deviation is small so the performance of the proposed method is quite consistent in every run. Therefore, we show the AUPRC curve of one of the 10 runs in Figure \ref{fig:thypr}.

\begin{table*}[ht]
\centering
\begin{tabular}{|c|c|c|c|c|} \hline
\multirow{2}{*}{Run} & \multicolumn{2}{c|}{AUPRC}& \multicolumn{2}{c|}{AUROC} \\
\cline{2-5}
 & Proposed Method & Isolation Forest &  Proposed Method & Isolation Forest \\ \hline
1 & 0.2822 & 0.1876 & 0.8952 & 0.9524 \\ \hline
2 &0.2327 & 0.143 &  0.8851 & 0.9464\\ \hline
3 &0.2417 & 0.1192 & 0.9075 & 0.9481\\ \hline
4 &0.3584 & 0.1097 & 0.9244 & 0.9485\\ \hline
5 &0.2743 & 0.1308 & 0.8498 & 0.9477\\ \hline
6 &0.2916 & 0.1959 & 0.9311 & 0.9523\\ \hline
7 &0.2269 & 0.1528 & 0.9194 & 0.9485\\ \hline
8 &0.2439 & 0.1232 & 0.9154 & 0.9411\\ \hline
9 &0.2729 & 0.102 & 0.8200 & 0.9492\\ \hline
10 &0.2319 & 0.1176 & 0.8892 & 0.9487\\ \hline
\end{tabular}
\caption{Performance on the Credit card Fraud Detection Dataset}\label{tab:auprc}
\end{table*}

\begin{table*}[ht]
\centering
\begin{tabular}{|c|c|c|c|c|} \hline
\multirow{2}{*}{Method} & \multicolumn{2}{c|}{AUPRC}& \multicolumn{2}{c|}{AUROC} \\
\cline{2-5}
 & Mean & Standard Deviation &  Mean & Standard Deviation \\ \hline
Proposed Method & 0.2656 & 0.0380 & 0.8937 & 0.0333 \\ \hline
Isolation Forest & 0.1381 & 0.0303 &  0.9482 & 0.0029\\ \hline
\end{tabular}
\caption{Mean and Standard Deviation of AUPRC and AUROC on credit card fraud detection dataset}\label{tab:meanstd}
\end{table*}

\subsection{Credit Card Fraud Detection}
In this section we do a thorough analysis of the proposed approach on the credit card fraud detection dataset available in Kaggle\cite{dal2015calibrating}. The dataset contains credit card transactions made in September 2013 by European cardholders in two days. It has 284,807 samples of which 492 samples are fraudulent transactions. The dataset is highly unbalanced as the positive class samples are only 0.172\% of all data points. This dataset is anonymized and contains only numerical input variables which are the result of a PCA transformation. It has 30 features out of which the only features which have not been transformed with PCA are \textit{Time} and \textit{Amount} features. \textbf{Time} is the seconds elapsed between each transaction and the first transaction in the dataset and \textbf{Amount} is the transaction amount. \textbf{Class} is the response variable and it takes value 1 in case of fraud and 0 otherwise.

Like previous experiments, we run our proposed algorithm 10 times and the results on AUPRC are shown in the Figure \ref{fig:prpm}. Note that the precision is high till recall of 0.4 and it drops after that. Therefore, we can identify 40\% of the fraud cases with the proposed algorithm with high precision. This is very useful in industry setting where a high percentage of fraudulent transactions are detected without any manual judgement.  

We compare our performance with the work of Pozzolo et al.\cite{dal2015calibrating}. They argued that for an extreme case of class imbalance such as this dataset, under-sampling can significantly improve the performance of the outlier detection algorithms. They showed the performance of Logit Boost, SVM and Random Forest for different values for the ratio of prior probabilities of two classes. Like most of the work in literature, they also chose AUROC as their metric and the results can be seen in the Figure 5 of their work\cite{dal2015calibrating}. For all three base methods, the AUC is above 0.9. Note that like  \cite{satimage2} and \cite{aggarwal2015theoretical}, Pozzolo et al.\cite{dal2015calibrating} also do not report numbers of their performance and only provide figures. Therefore we could only do visual comparison. 

\begin{figure*}
\centering
\includegraphics[width=.15\textwidth]{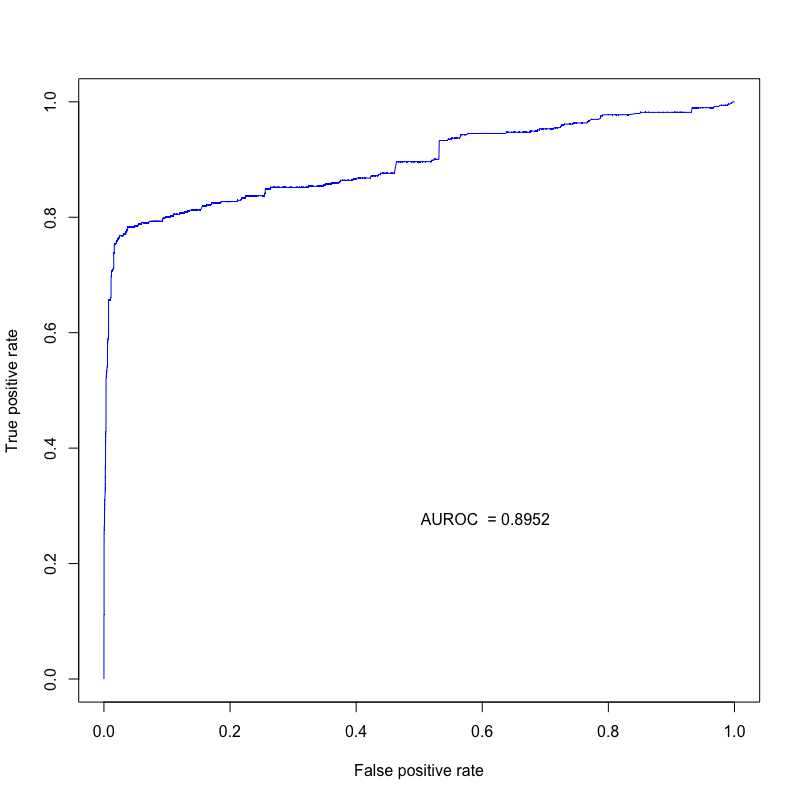}\quad
\includegraphics[width=.15\textwidth]{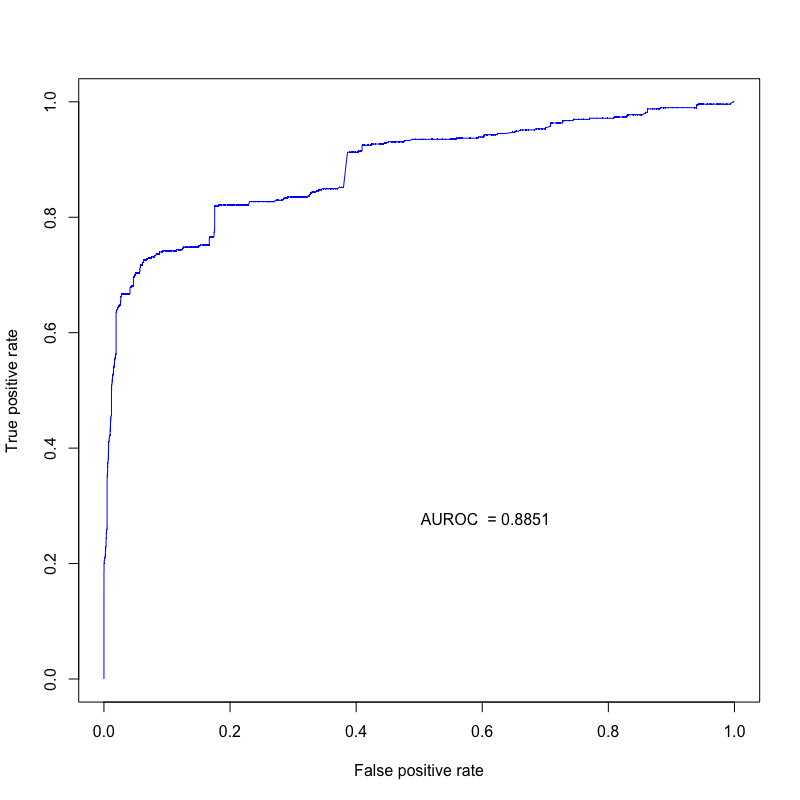}\quad
\includegraphics[width=.15\textwidth]{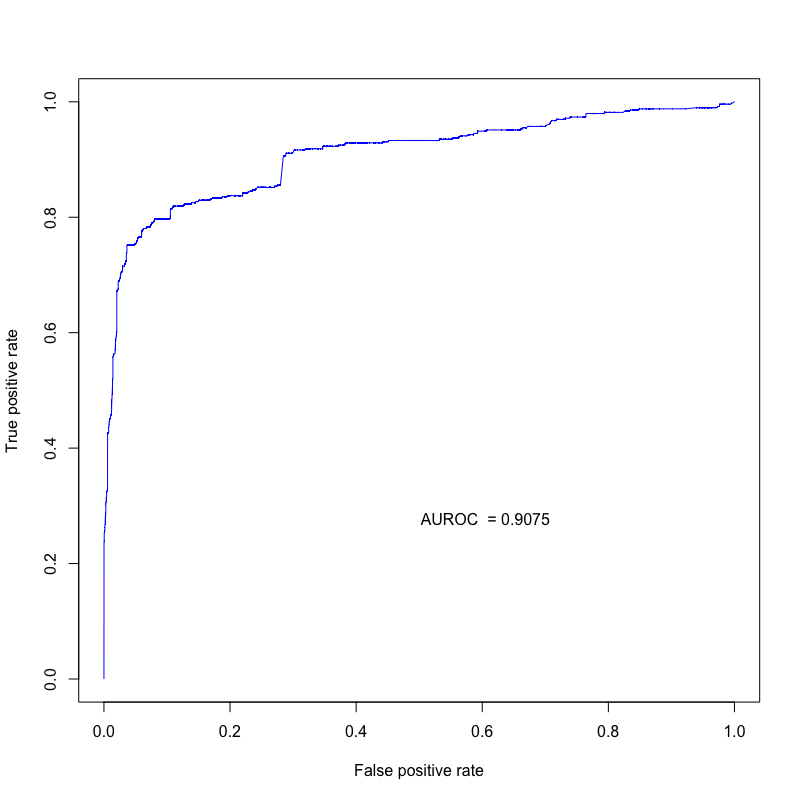}\quad
\includegraphics[width=.15\textwidth]{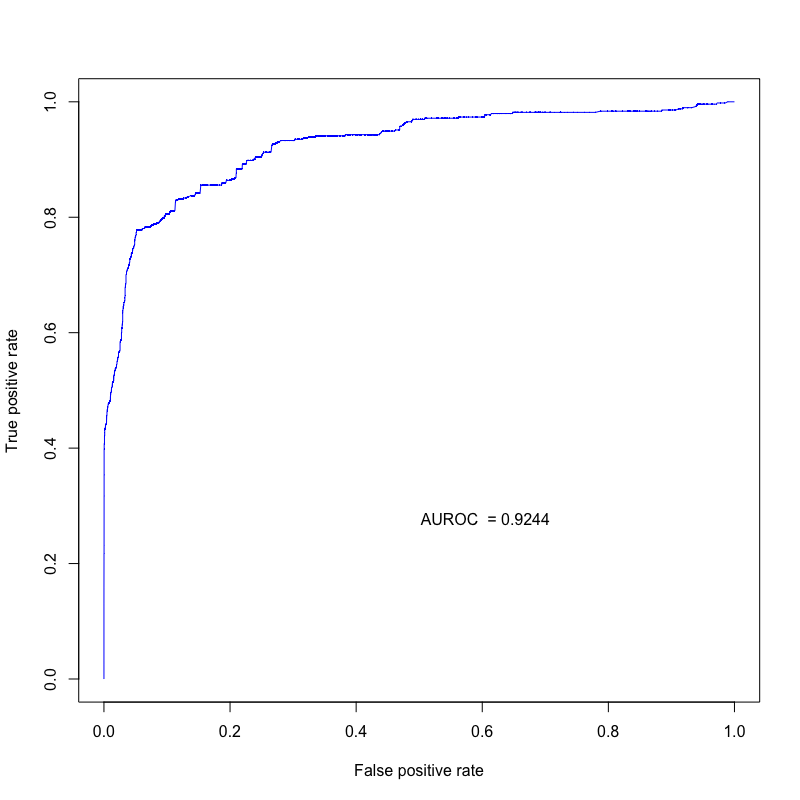}\quad
\includegraphics[width=.15\textwidth]{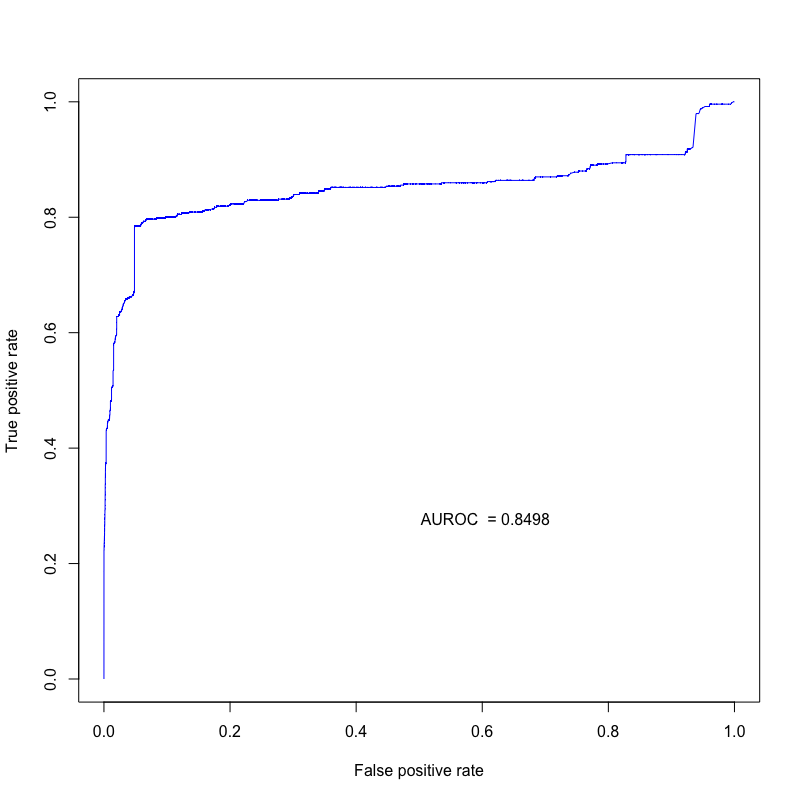}

\medskip

\includegraphics[width=.15\textwidth]{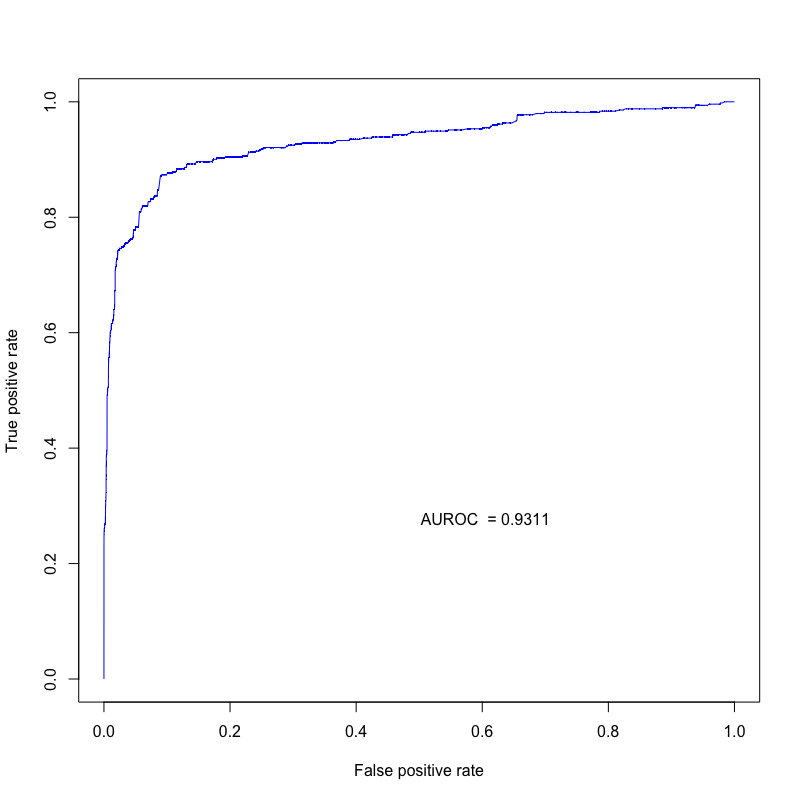}\quad
\includegraphics[width=.15\textwidth]{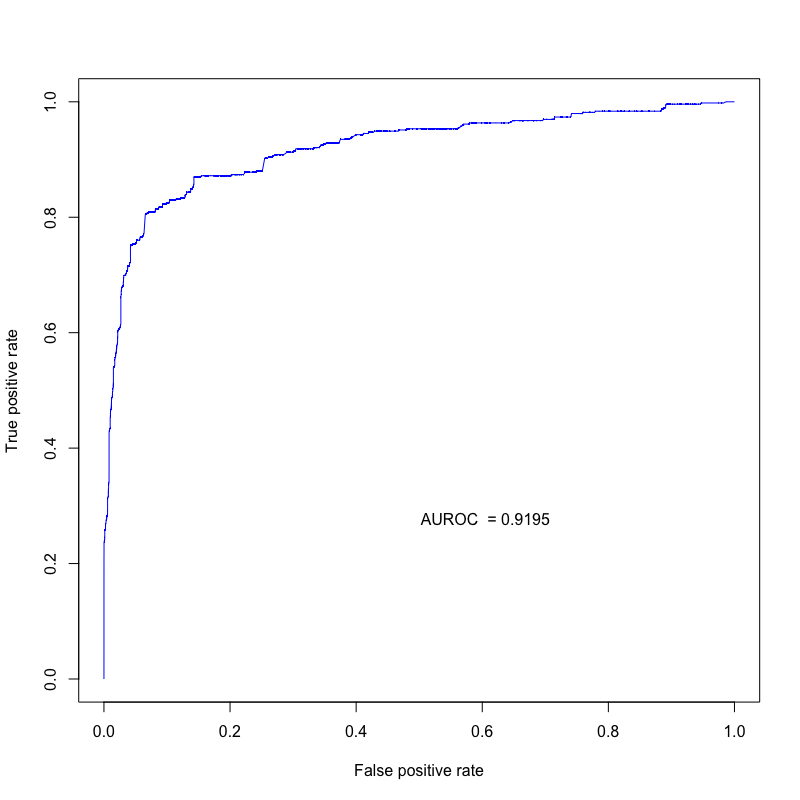}\quad
\includegraphics[width=.15\textwidth]{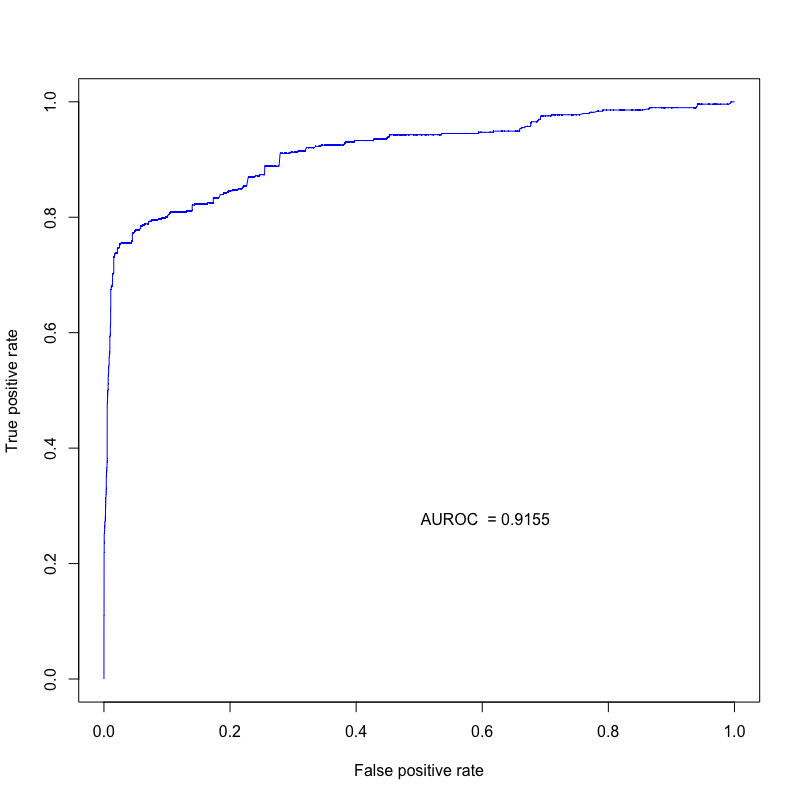}\quad
\includegraphics[width=.15\textwidth]{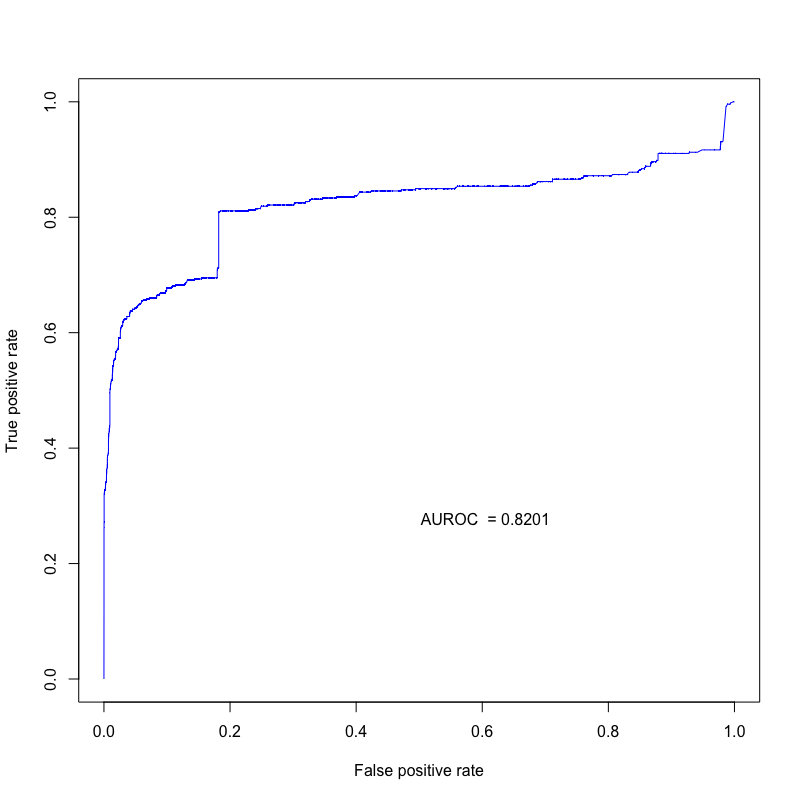}\quad
\includegraphics[width=.15\textwidth]{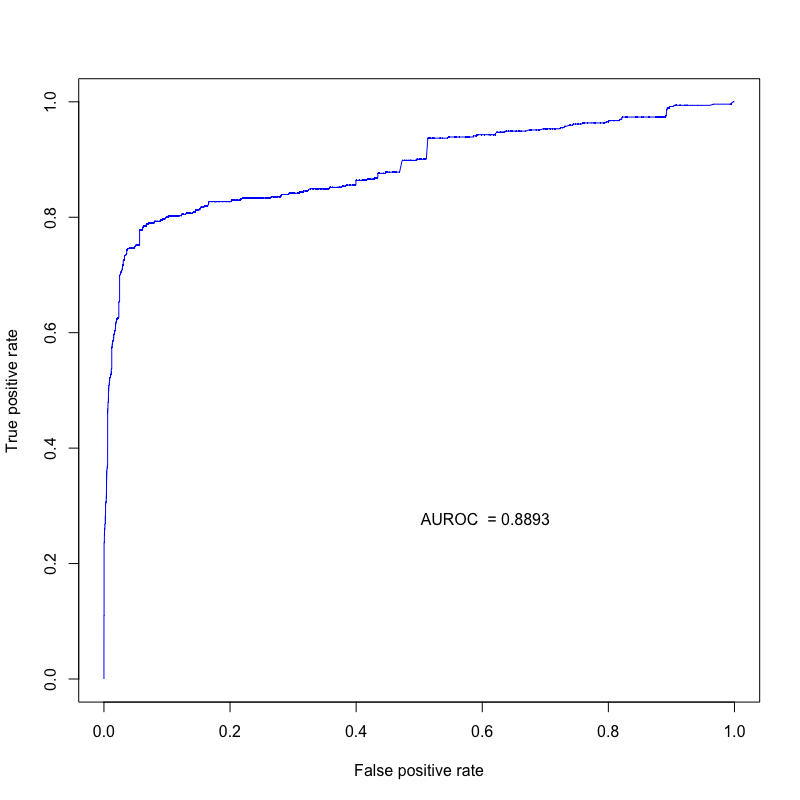}

\caption{ROC curves of different runs of the proposed algorithm}
\label{fig:rocpm}
\end{figure*}

\begin{figure*}
\centering
\includegraphics[width=.15\textwidth]{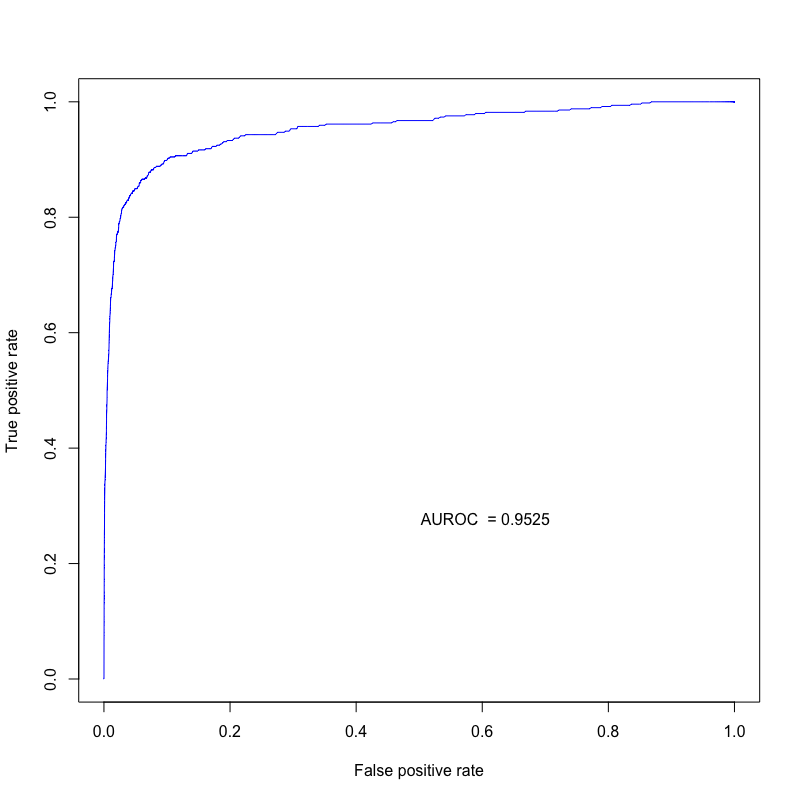}\quad
\includegraphics[width=.15\textwidth]{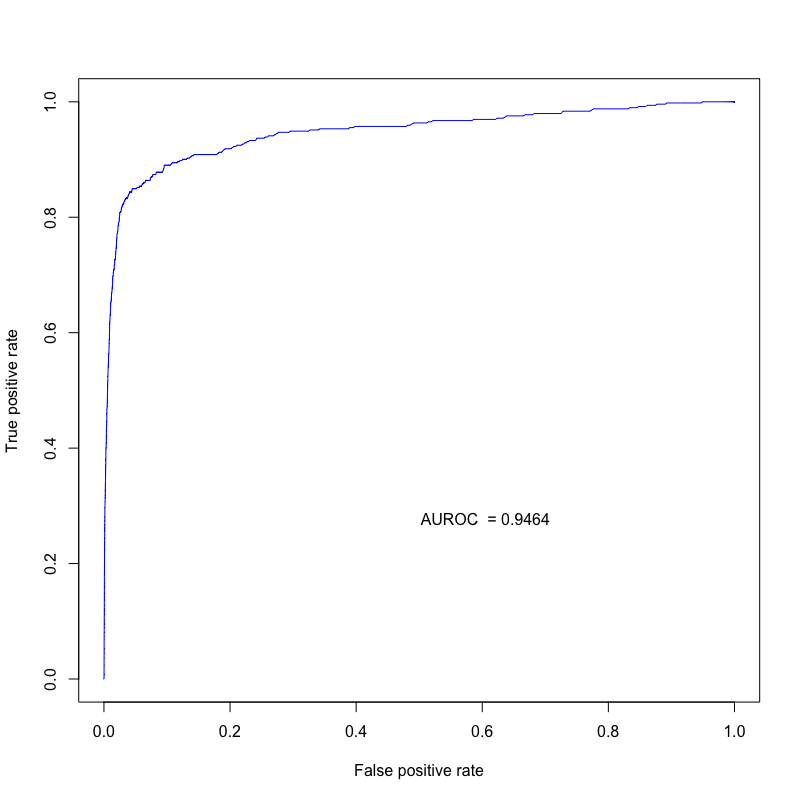}\quad
\includegraphics[width=.15\textwidth]{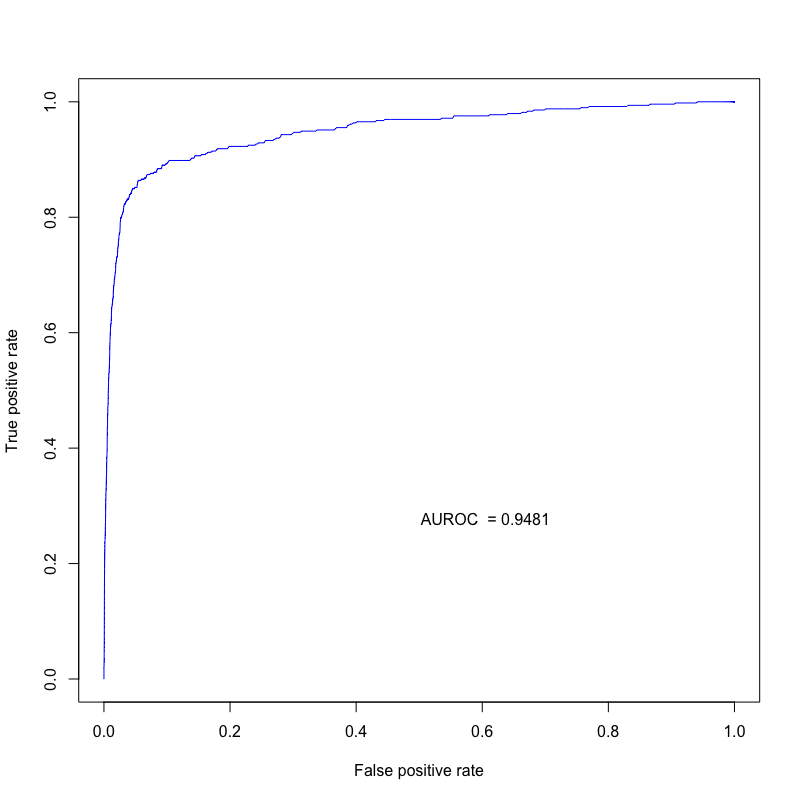}\quad
\includegraphics[width=.15\textwidth]{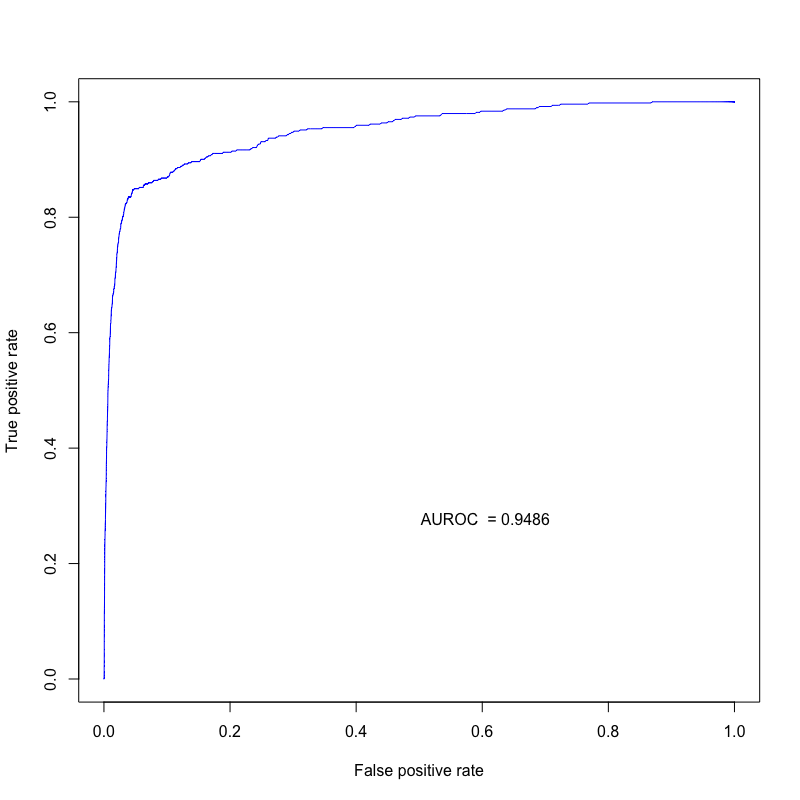}\quad
\includegraphics[width=.15\textwidth]{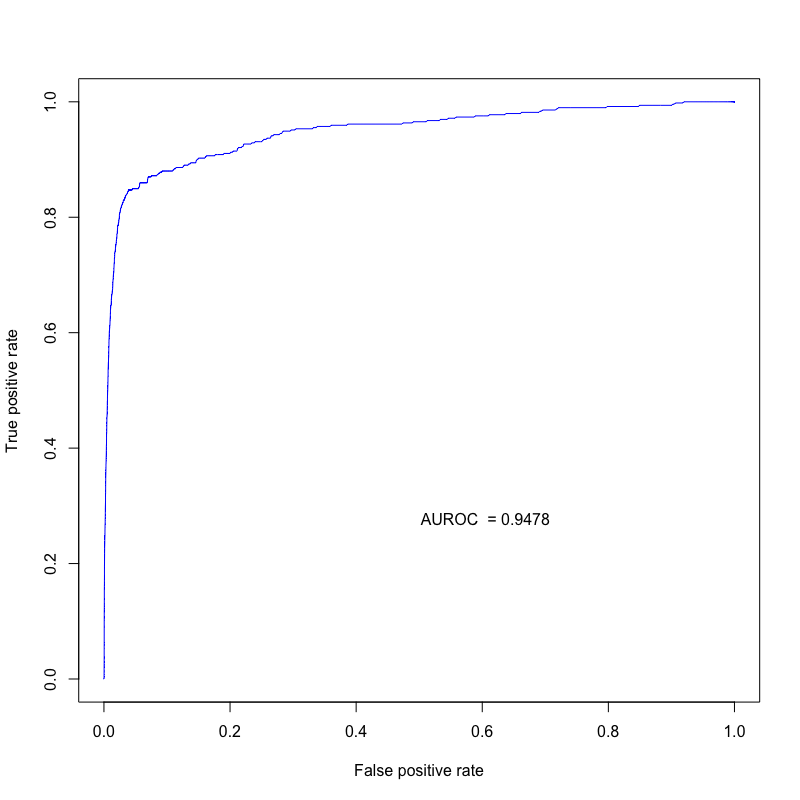}

\medskip

\includegraphics[width=.15\textwidth]{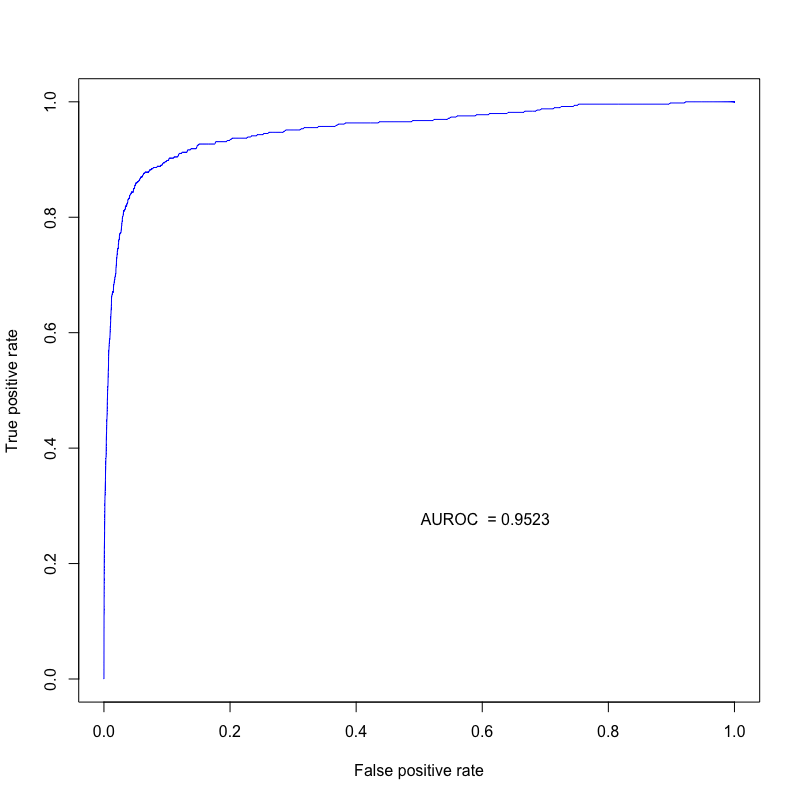}\quad
\includegraphics[width=.15\textwidth]{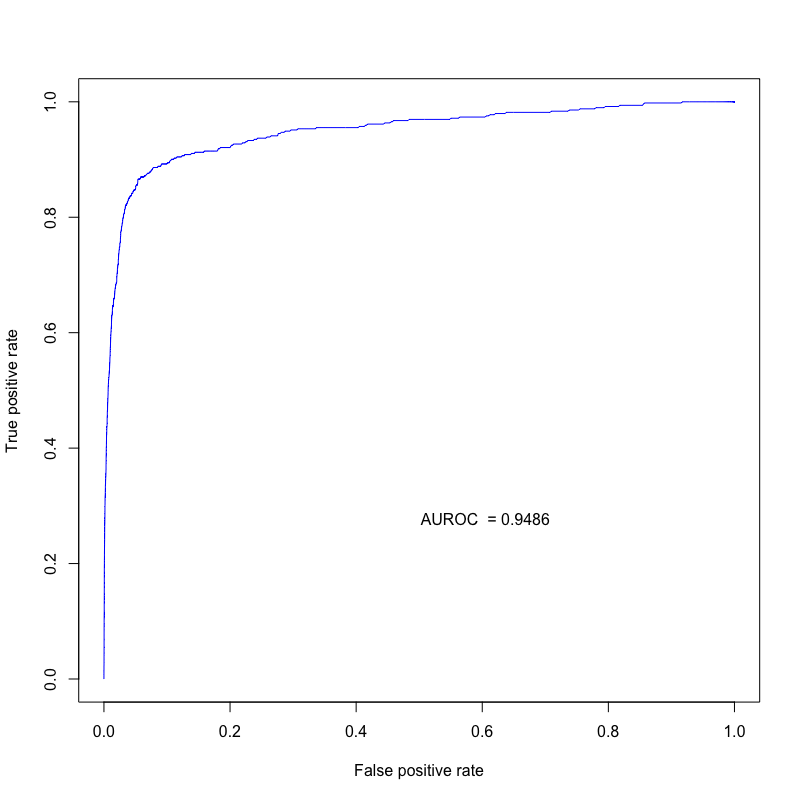}\quad
\includegraphics[width=.15\textwidth]{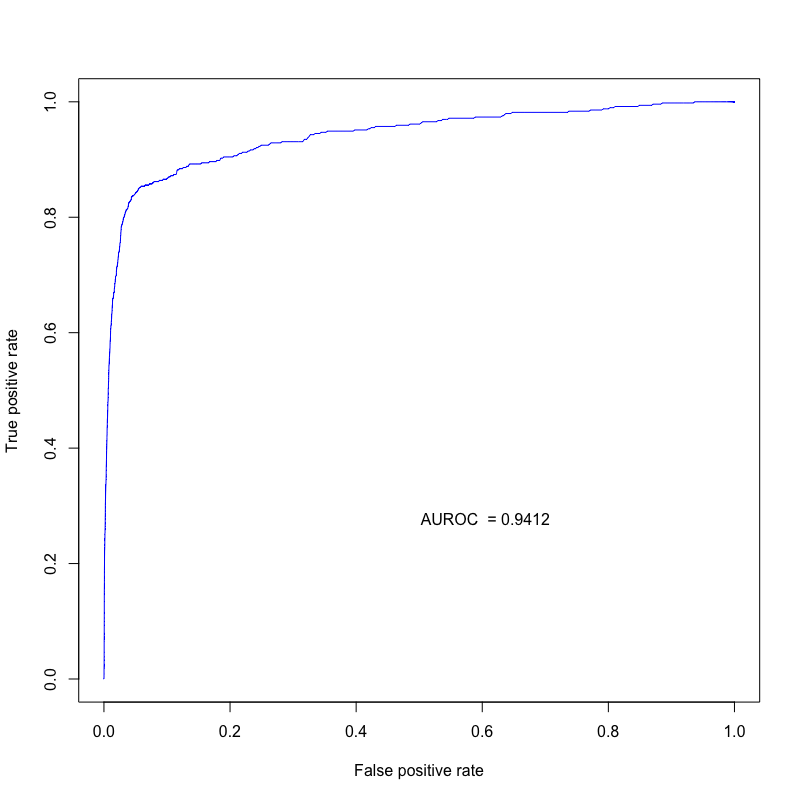}\quad
\includegraphics[width=.15\textwidth]{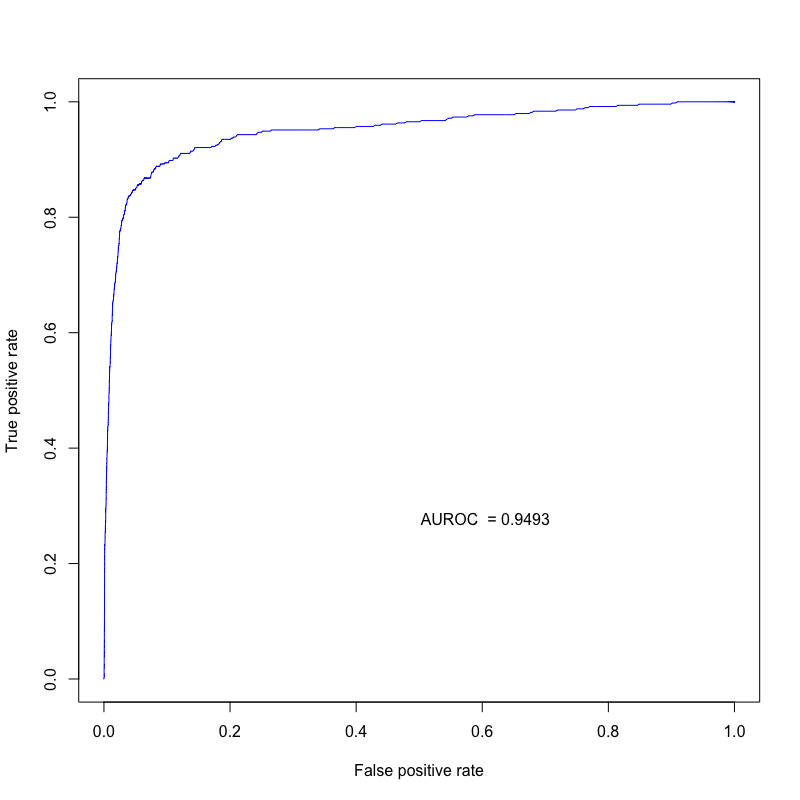}\quad
\includegraphics[width=.15\textwidth]{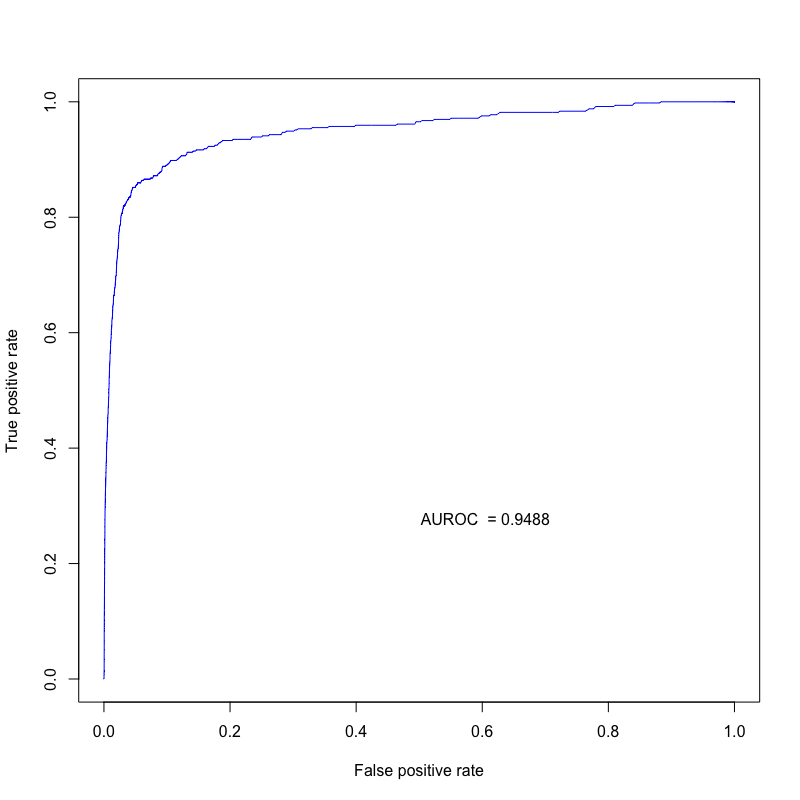}

\caption{ROC curves of different runs of Isolation Forest}
\label{fig:rocif}
\end{figure*}

We also calculated the AUROC for our method and the results are shown in the Figure \ref{fig:rocpm} and Tables \ref{tab:auprc} and \ref{tab:meanstd}. Our mean AUROC is 0.8937 with a standard deviation of 0.033. However, as explained earlier, we believe for such an imbalance dataset, AUROC presents a misleading view of the performance of the outlier detection algorithm. To validate our claim, we also ran Isolation Forest\cite{liu2008isolation} on the credit card fraud detection dataset. Isolation Forest builds a tree like structure by randomly selecting and then splitting feature to isolate every data point. Authors argue that the outliers are often closer to the root of the tree while the inliers are isolated at the deeper end. Isolation Forest gets comparable performance to the work of Pozzolo et al.\cite{dal2015calibrating}  as shown in Figure \ref{fig:rocif} and Tables \ref{tab:auprc} and \ref{tab:meanstd} but perform significantly worse than our proposed method on AUPRC as showed in the Figure \ref{fig:prif} and Tables \ref{tab:auprc} and \ref{tab:meanstd}. This validates our claim that to get an accurate estimate of the performance of the outlier detection algorithm we should use AUPRC as the metric. It is possible to get good AUROC because of good performance on true negatives (non-fraudulent transactions) which is not as important as getting the true positives correct in a use case like credit card fraud detection. This is exhibited by Isolation Forest where it outperforms our proposed method in terms of AUROC but does significantly worse than the proposed algorithm on AUPRC. 

\begin{figure*}
\centering
\includegraphics[width=.3\textwidth]{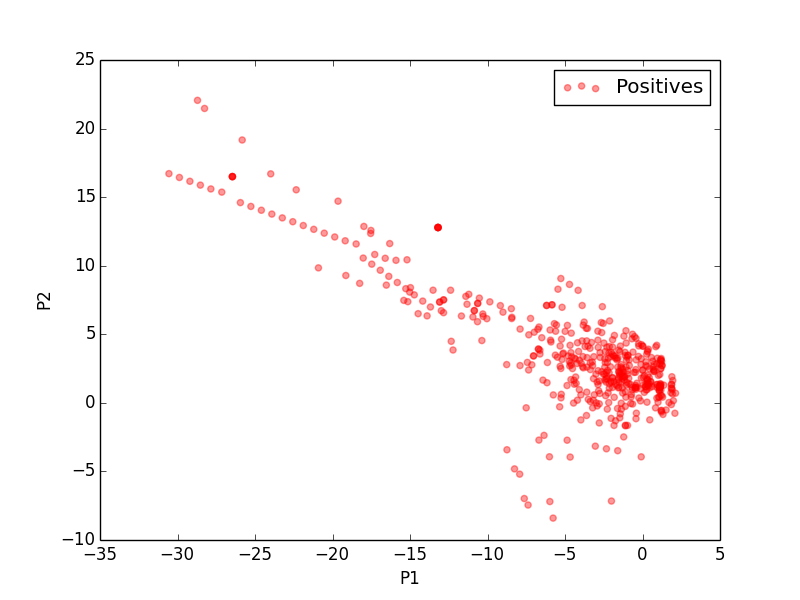}\quad
\includegraphics[width=.3\textwidth]{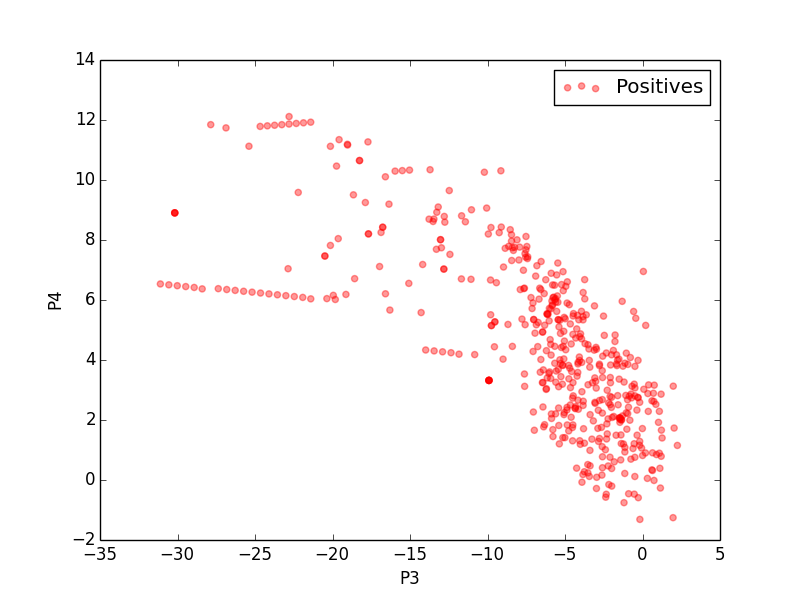}\quad
\includegraphics[width=.3\textwidth]{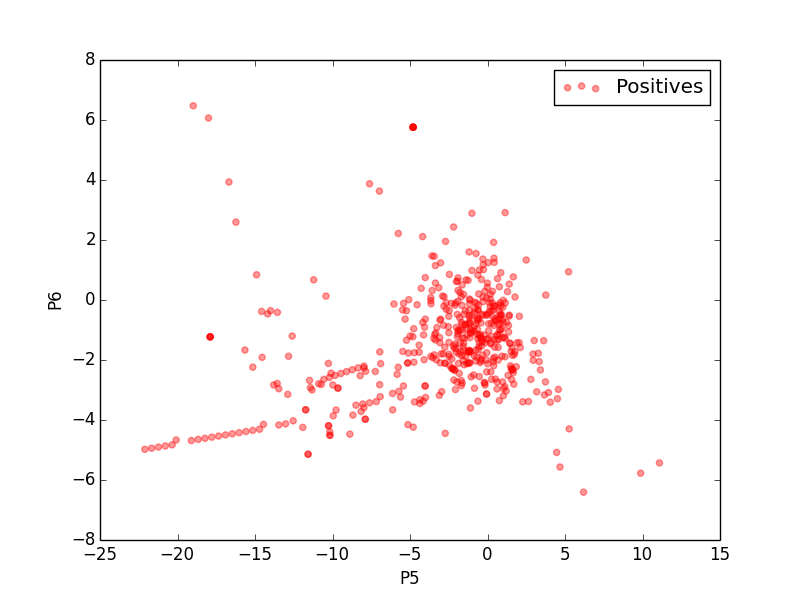}


\includegraphics[width=.3\textwidth]{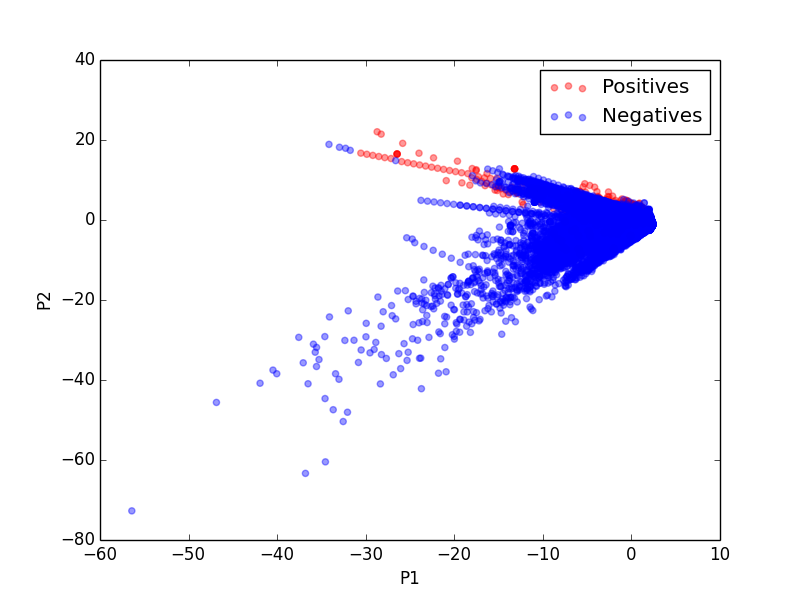}\quad
\includegraphics[width=.3\textwidth]{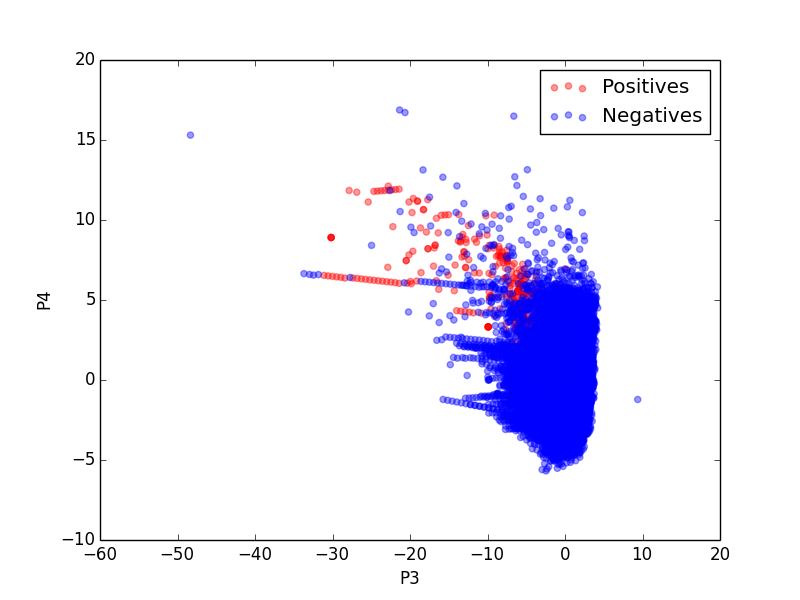}\quad
\includegraphics[width=.3\textwidth]{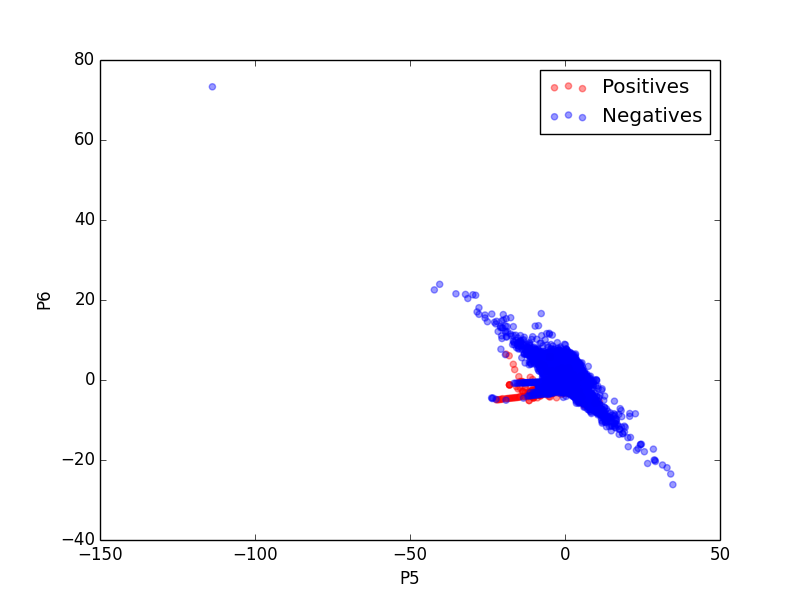}


\includegraphics[width=.3\textwidth]{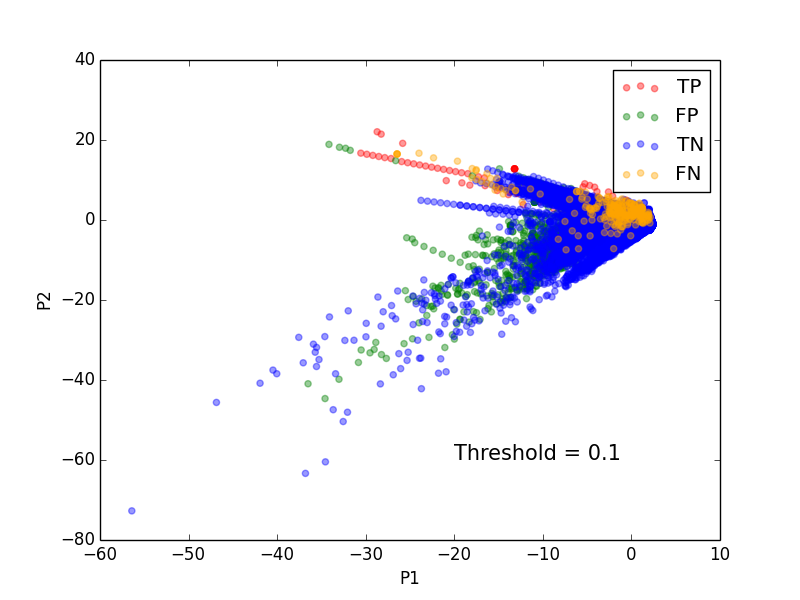}\quad
\includegraphics[width=.3\textwidth]{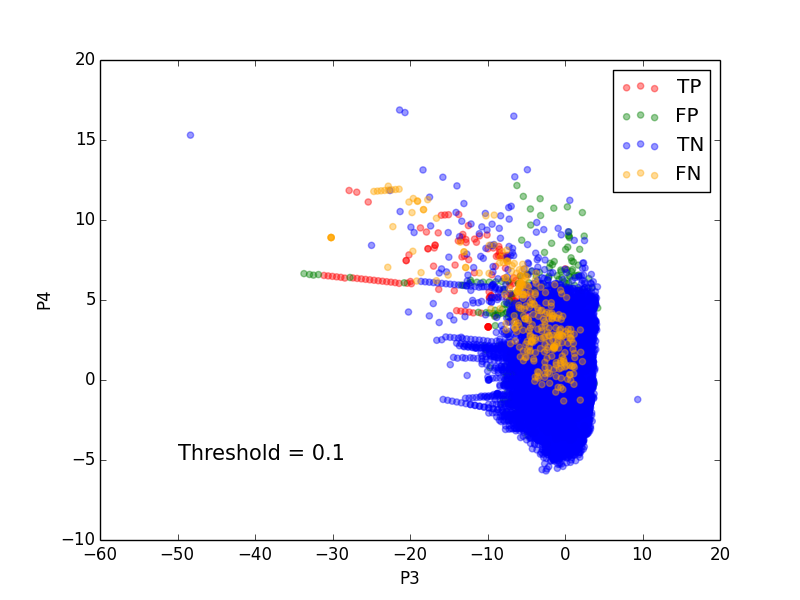}\quad
\includegraphics[width=.3\textwidth]{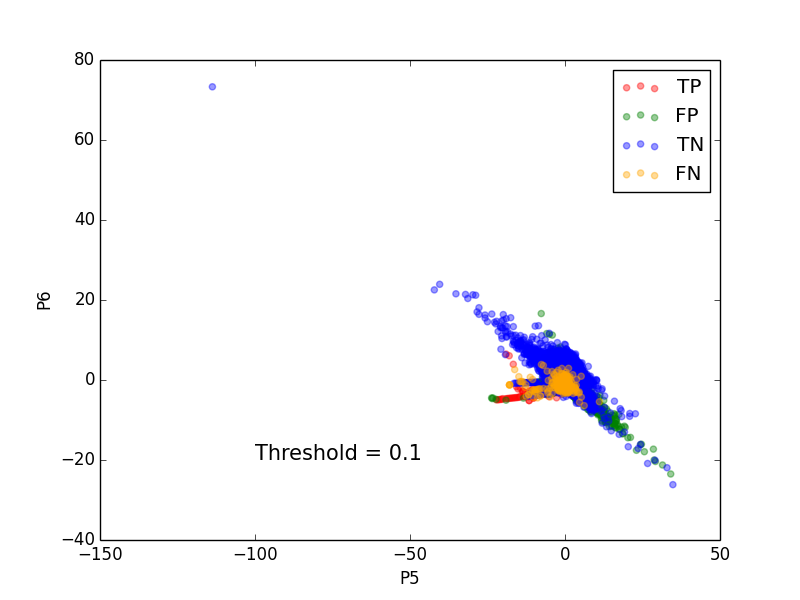}


\includegraphics[width=.3\textwidth]{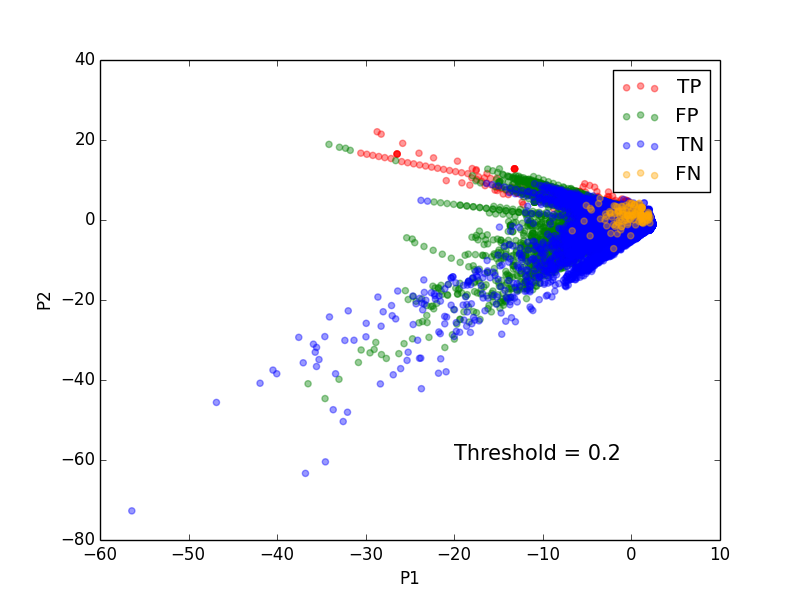}\quad
\includegraphics[width=.3\textwidth]{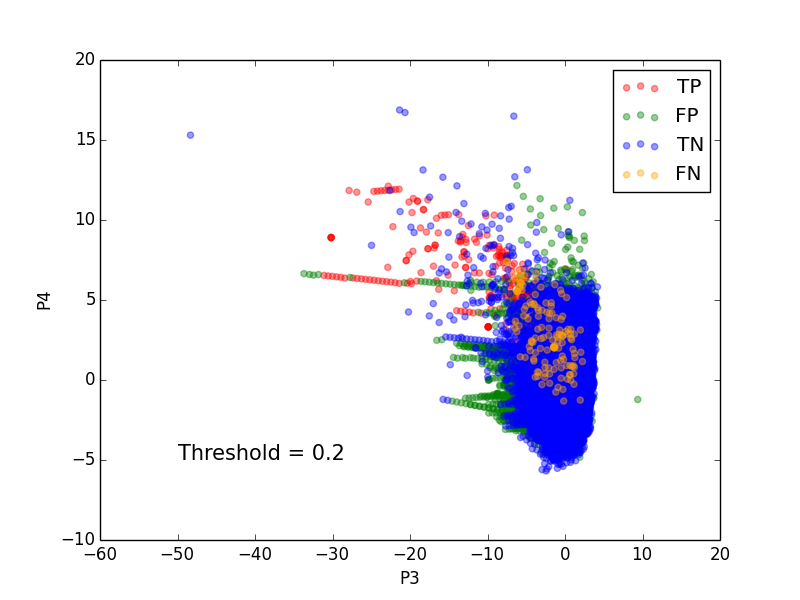}\quad
\includegraphics[width=.3\textwidth]{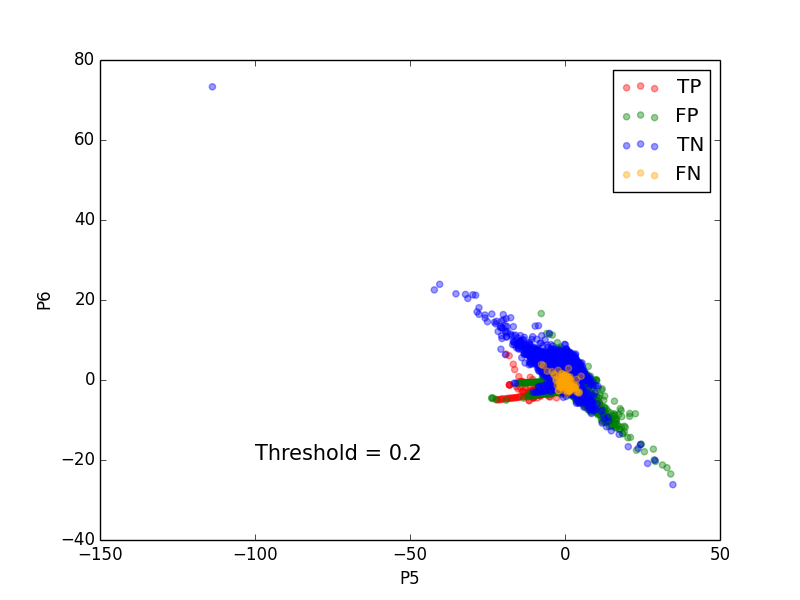}


\includegraphics[width=.3\textwidth]{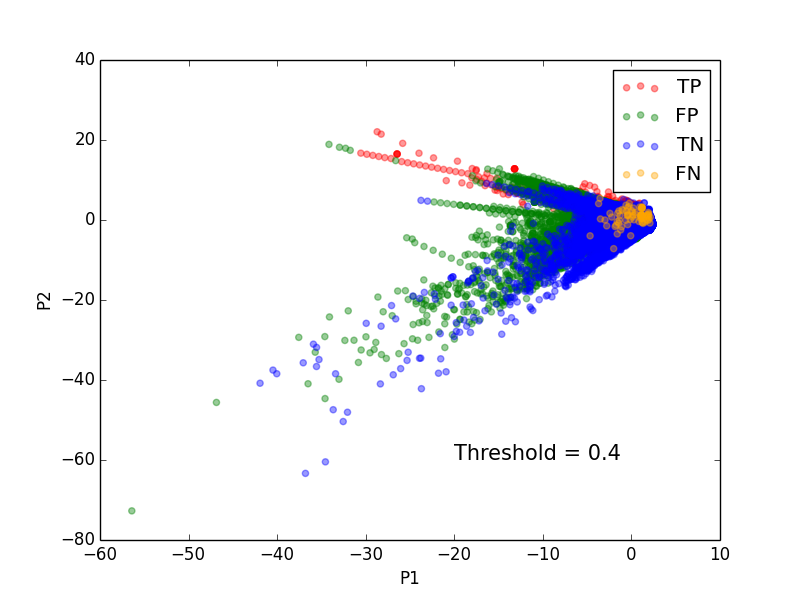}\quad
\includegraphics[width=.3\textwidth]{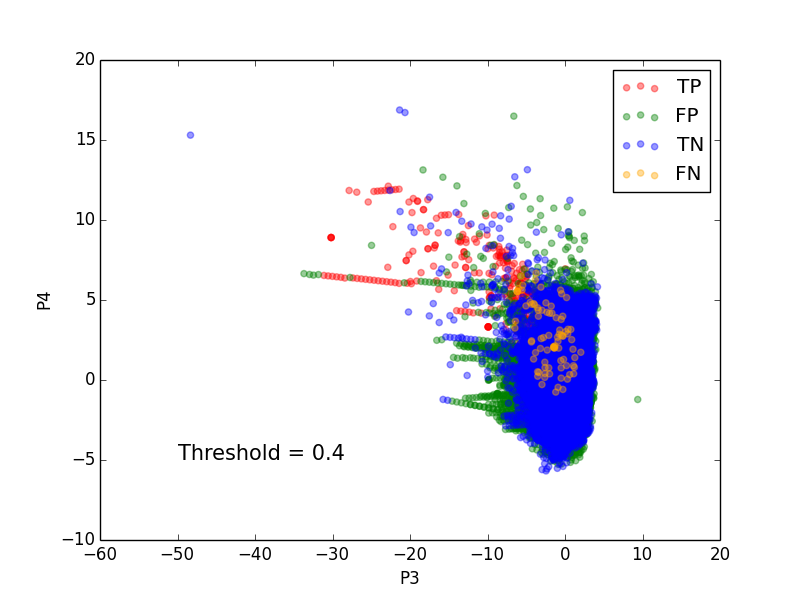}\quad
\includegraphics[width=.3\textwidth]{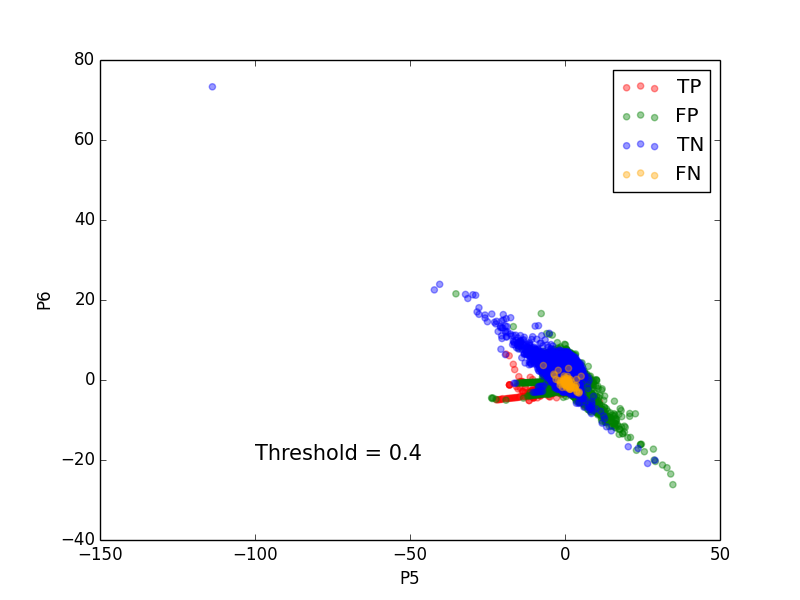}


\includegraphics[width=.3\textwidth]{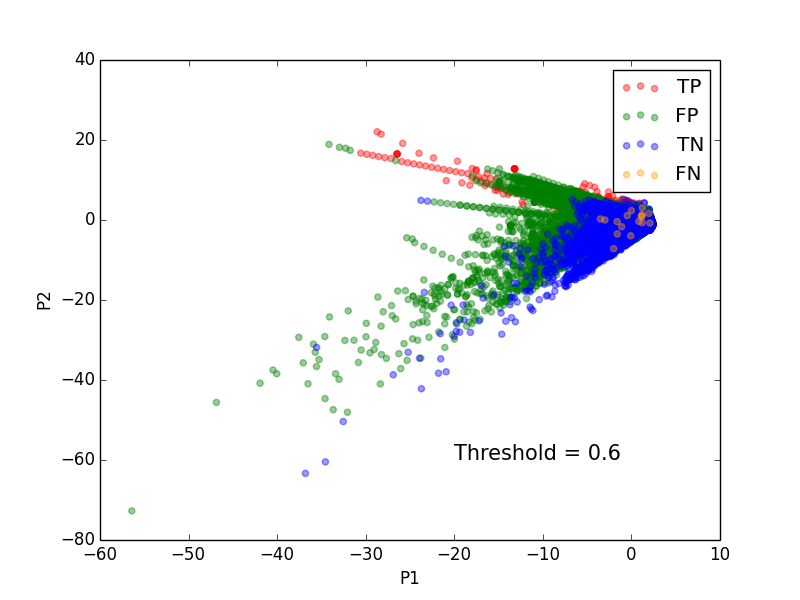}\quad
\includegraphics[width=.3\textwidth]{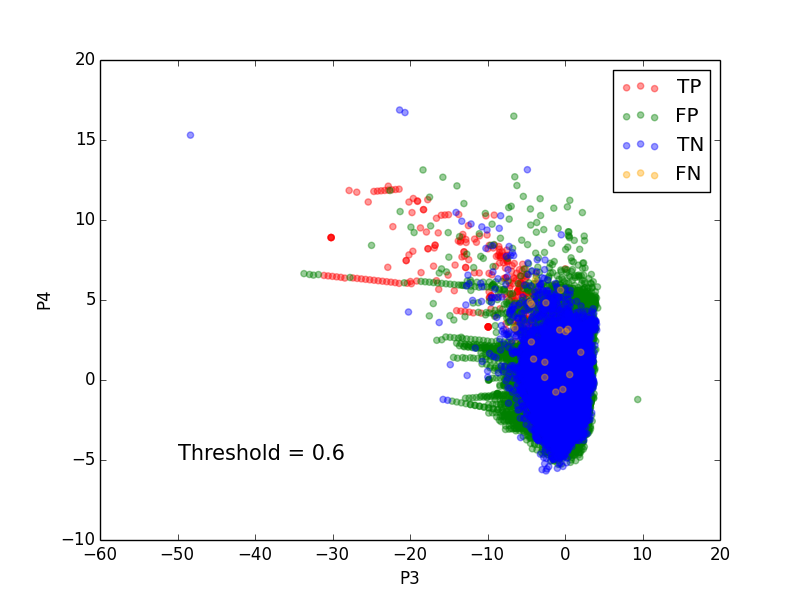}\quad
\includegraphics[width=.3\textwidth]{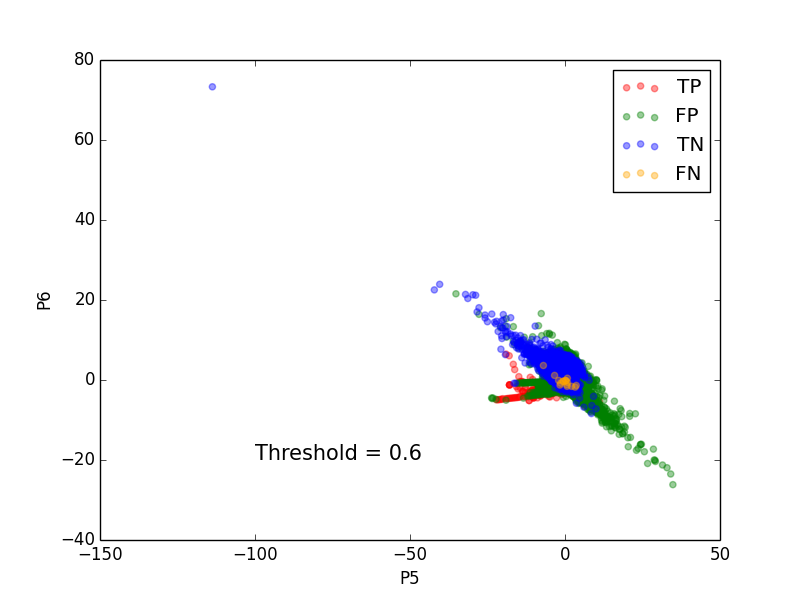}

%
\caption{Scatter plots of different views of the credit card fraud detection dataset along different principal components}
\label{fig:scatter}
\end{figure*}

Precision Recall curves of different runs of Isolation Forest on the credit card detection dataset is shown in the Figure \ref{fig:prif}. Note that Isolation Forest draws a fixed number of samples to train each base estimator. Therefore, running it different times will result in different results because of the different samples drawn each time. To account for this behavior, we ran Isolation Forest 10 times just like the proposed method. Since, \textit{n}th run of the proposed method is run on different dataset than the \textit{n}th run of Isolation Forest we cannot plot the performance of two algorithm together for each run. Therefore, we have plotted their 10 runs separately as it is not possible to compare the two simultaneously. 

We also conducted an experiment to address the issue of \textit{k}-means to the order in which data is presented. In credit card fraud detection dataset, one of the non-transformed features is time. Each data point has a time feature which is the time elapsed between that transaction and the first transaction. We presented the data using time i.e. as the transaction occurred and removed time from further analysis. We get an AUPRC of 0.2231. This performance of the proposed method is significant because in AUROC, there is an implicit baseline value of 0.5 which represents when the model has no information or predictive power. In AUPRC on the other hand, there is no such implicit baseline value. So, a baseline algorithm with no predictive power will flag everything as an outlier and it'll be right 0.172\% times (number of positive samples in the dataset). Therefore, in our dataset a baseline model with no predictive power will have an AUPRC of 0.00172. As shown in the Table \ref{tab:auprc} we get a mean AUPRC of 0.2656 with minimal standard deviation and for time ordered experiment we get an AUPRC 0.2231. In both cases it is a significant jump from the baseline model of no predictive power. 

\section{Observations}
\label{sec:observation}

Since the credit card fraud detection dataset is transformed using PCA except for time and amount features, it is safe to assume that the individual dimensions are orthogonal to each other. We estimated variance captured by each dimension as a proxy for the information contained in the respective dimensions. As shown in Table \ref{tab:pcvar} first six principal components (out of 28) contains about half of the information (47.31\% of variance). So, we decided to plot the first principal component against second, third against fourth and fifth and sixth to get different views of the data. We plotted original data along with how our proposed algorithm does at different thresholds and is shown in the Figure \ref{fig:scatter}. The second row is of the original data showing how outliers are placed against inliers. As it can be seen that a good chunk of outlier are separated from inliers as evident in the plot of first principal component against second. However, there are some fraudulent transaction that overlap with non-fraudulent transactions in all three views of the data. These data points are difficult to visualize  as they overlap so positive ones are hidden behind the negative ones. To visualize that positive samples are there hidden behind the negative ones, we plotted only positive samples in the first row. Since we plotted only the positive samples, the plots in the first row are in a different scale than the plots in the rest of rows and they appear to be zoomed in. However, it is clear by looking into the plots in row 1 and 2 that there are overlapping positive and negative samples and only negative ones are shown in row 2. 

It is clear that separating the outliers from inliers when they are overlapping will be really challenging for any algorithm. As shown in the third row of the Figure \ref{fig:scatter}, if we go with a threshold of 0.1 where we flag every data point with less than 0.1 score as outliers we get lot of true positives right with some false positives where outliers are separated from inliers. However, as expected algorithm has difficulty separating inliers from outliers when they overlap. Therefore, we see a lot of false negatives. In this case an ideal method would be the one that can maximize the true positives while minimizing the false negatives. As we increase the threshold we see the number of false negatives going down but number of false positive keep on increasing. The right trade-off in this situation will depend on the application and domain. For instance, in situations where only fixed number of recommendations can only be made for human judgements then we would want those recommendations to be high quality. Therefore, our goal would be to maximize precision in that case and choose a low threshold. 

As a concrete example consider a use case where the team responsible for detecting fraudulent transactions has to make recommendation to the customer service team of the most likely fraudulent transactions that they can examine. In such use case we can only make limited recommendation due to limited resources and limitations of human judgement. We can observe that in Figure \ref{fig:pr} for a recall of 0.4, we have a precision of 0.1. Therefore, we can detect 40\% of the 492 (~200) outliers by manually examine only about 2000 data points. This is a huge gain as the total number of data points are 284,807 data points. We are getting great returns on the recommendations made using the proposed method. Different threshold can be chosen according to the requirement and capacity of the team.
\begin{table*}[ht]
\centering
\begin{tabular}{|c|c|c|c|c|c|} \hline
\multirow{2}{*}{} & \multicolumn{5}{c|}{Threshold} \\
\cline{2-6}
 & >0.5 & >0.6 &  >0.7 & >0.8 & 0.9 \\ \hline
True Negatives & 139220 & 72462 & 29071 & 20201 & 17336 \\ \hline
True Positives & 29 & 17 & 12 & 6 & 5 \\ \hline
\end{tabular}
\caption{Number of true negatives and true positives in the higher range of consistency scores on credit card fraud detection dataset ordered by time of the transactions}\label{tab:novelty}
\end{table*}

\begin{table}[ht]
\centering
\begin{tabular}{|c|c|} \hline
Principal component & Percentage of Variance \\ \hline
1 & 12.48 \\ \hline
2 & 8.87 \\ \hline
3 & 7.48 \\ \hline 
4 & 6.52 \\ \hline 
5 & 6.19 \\ \hline 
6 & 5.77 \\ \hline
7 & 4.97 \\ \hline 
8 & 4.64 \\ \hline 
9 & 3.92 \\ \hline 
10 & 3.85 \\ \hline 
11 & 3.39 \\ \hline 
12 & 3.24 \\ \hline
13 & 3.22 \\ \hline 
14 & 2.99 \\ \hline 
15 & 2.72 \\ \hline 
16 & 2.49 \\ \hline 
17 & 2.34 \\ \hline 
18 & 2.28 \\ \hline
19 & 2.15 \\ \hline 
20 & 1.93 \\ \hline 
21 & 1.75 \\ \hline 
22 & 1.71 \\ \hline 
23 & 1.26 \\ \hline 
24 & 1.19 \\ \hline
25 & 0.88 \\ \hline
26 & 0.75 \\ \hline 
27 & 0.53 \\ \hline 
28 & 0.35 \\ \hline
\end{tabular}
\caption{Percentage of variance captured by different principal components of credit card fraud detection dataset}\label{tab:pcvar}
\end{table}

Another use case would be if the team wants to build a training set for a novelty detection system or for learning a one-class classifier. One way in which novelty detection differs from outlier detection is that in novelty detection the training dataset should not be contaminated. That is in ideal condition it should not have any anomalies or outliers. Our proposed method can be used to collect such \textit{pure} dataset by selecting the data samples with high consistency score. In the credit card detection dataset we can observe that data points with high consistency score were largely true negatives as shown in Table \ref{tab:novelty}. Such training set can be used to learn good consumer behavior.   

As can be observed from Table \ref{tab:novelty} our proposed method is highly accurate in identifying true negatives or good consumer behavior. Our method can be immensely helpful as out of  284,807 samples we can safely rule out 139220 as true negatives based on threshold of 0.5. We will make few mistakes (29) that is acceptable in industry at this scale. Moreover, we can also identify about 40\% of the false positives with recommending only about 2000 data samples. 




\section{Conclusion}
\label{sec:conclusion}
In this paper we propose a method that shows tremendous potential in identifying outliers and pure inliers by assigning a consistency score to each data point.  The proposed method assumes no prior knowledge of either the outliers or inliers. We showed that application of proposed method in different scenarios such as to make recommendation for potential outliers for further investigation with high precision and to create training sets for novelty detection algorithms. We also argued for a better evaluation metric and showed that area under the precision recall curve is better than area under the ROC curve for outlier detection problems. Lastly, we showed the efficacy of our method on both UCI datasets and a real world credit card fraud detection dataset.

Formalization of the proposed approach is left as future work.

\bibliographystyle{ACM-Reference-Format}
\bibliography{sample-bibliography}

\end{document}